\documentclass[fleqn, 10pt]{wlscirep}
\usepackage[utf8]{inputenc}
\usepackage[T1]{fontenc}
\usepackage{colortbl}
\usepackage{bm}
\usepackage{marvosym}
\usepackage{multirow} 

\title{A Unified Low-level Foundation Model for Enhancing Pathology Image Quality}

\author[1]{Ziyi Liu}
\author[1]{Zhe Xu}
\author[1]{Jiabo Ma}
\author[1]{Wenqaing Li}
\author[1]{Junlin Hou}
\author[1]{Fuxiang Huang}
\author[1]{Xi Wang}
\author[2]{Ronald Cheong Kin Chan}
\author[3]{Terence Tsz Wai Wong}
\author[1,3,4,5,6 \Letter]{Hao Chen}

\affil[1]{Department of Computer Science Engineering, The Hong Kong University of Science and Technology, Hong Kong, China}
\affil[2]{Department of Anatomical and Cellular Pathology, The Chinese University of Hong Kong, Hong Kong, China}
\affil[3]{Department of Chemical and Biological Engineering, Hong Kong University of Science and Technology, Hong Kong, China}
\affil[4]{Division of Life Science, Hong Kong University of Science and Technology, Hong Kong, China}
\affil[5]{HKUST Shenzhen-Hong Kong Collaborative Innovation Research Institute, Futian, Shenzhen, China}
\affil[6]{State Key Laboratory of Nervous System Disorders, The Hong Kong University of Science and Technology, Hong Kong, China}

\affil[]{\textbf{Corresponding author: Hao Chen (jhc@cse.ust.hk)}}

\begin{abstract}
Foundation models have revolutionized computational pathology by achieving remarkable success in high-level diagnostic tasks, yet the critical challenge of low-level image enhancement remains largely unaddressed. Real-world pathology images frequently suffer from degradations such as noise, blur, and low resolution due to slide preparation artifacts, staining variability, and imaging constraints, while the reliance on physical staining introduces significant costs, delays, and inconsistency. Although existing methods target individual problems like denoising or super-resolution, their task-specific designs lack the versatility to handle the diverse low-level vision challenges encountered in practice. To bridge this gap, we propose the first unified Low-level Pathology Foundation Model (LPFM), capable of enhancing image quality in restoration tasks, including super-resolution, deblurring, and denoising, as well as facilitating image translation tasks like virtual staining (H\&E and special stains), all through a single adaptable architecture.Our approach introduces a contrastive pre-trained encoder that learns transferable, stain-invariant feature representations from 190 million unlabeled pathology images, enabling robust identification of degradation patterns. A unified conditional diffusion process dynamically adapts to specific tasks via textual prompts, ensuring precise control over output quality. Trained on a curated dataset of 87,810 whole slied images (WSIs) across 34 tissue types and 5 staining protocols, LPFM demonstrates statistically significant improvements (p<0.01) over state-of-the-art methods in most tasks (56/66), achieving  Peak Signal-to-Noise Ratio (PSNR) gains of 10-15\% for image restoration and Structural Similarity Index Measure (SSIM) improvements of 12–18\% for virtual staining. More importantly, LPFM represents a transformative advancement for digital pathology, as it not only overcomes fundamental image quality barriers but also establishes a new paradigm for stain-free, cost-effective, and standardized pathological analysis, which is crucial for enabling scalable and equitable deployment of AI-assisted pathology worldwide.
\end{abstract}

\begin{document}

    % \flushbottom
    \maketitle

\begin{figure*}
    \centering
    \includegraphics[width=\linewidth]{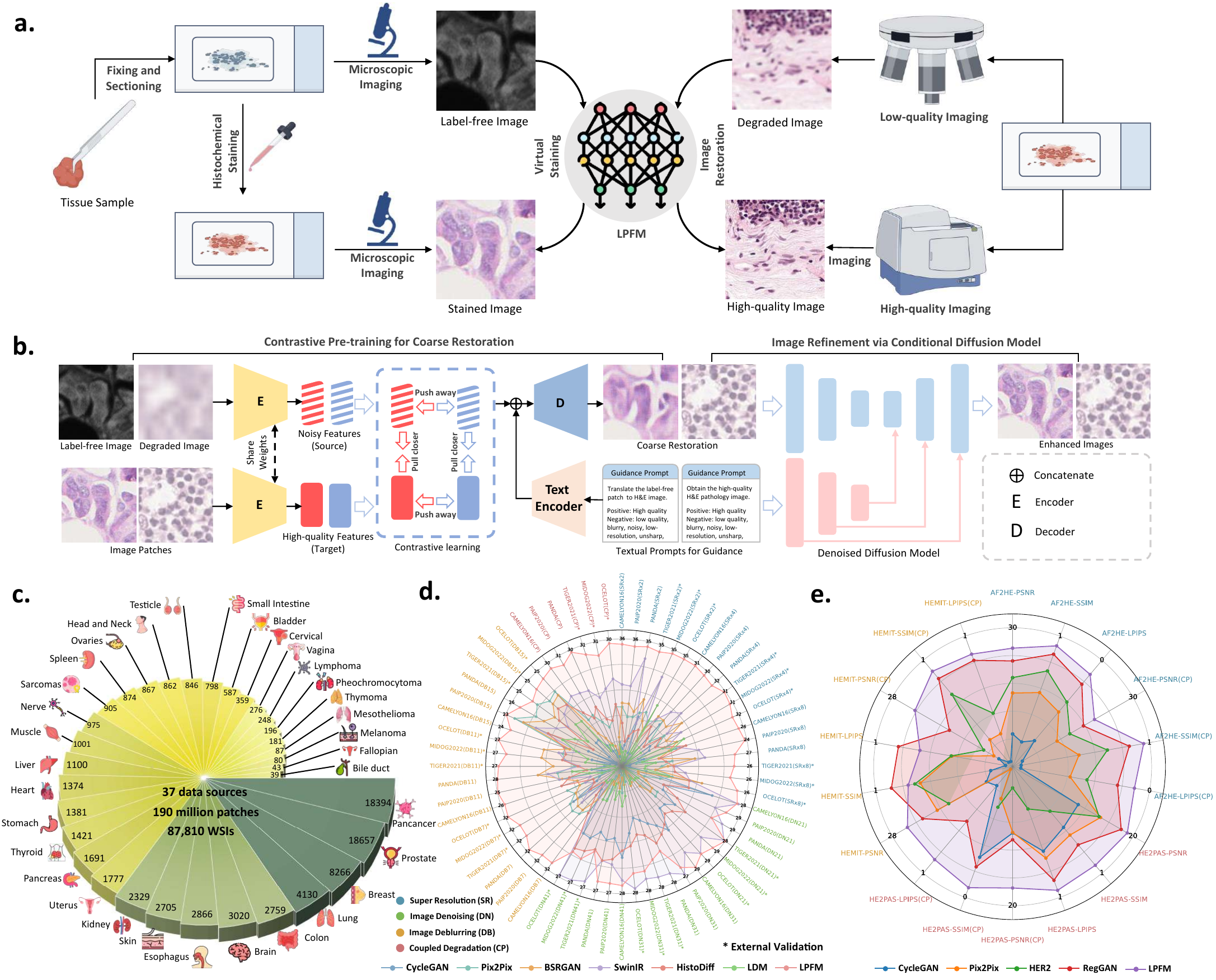}
    \caption{\textbf{Overview of the proposed low-level pathology foundation model (LPFM) for pathology image restoration and virutal staining.} 
    \textbf{a.} The pathology imaging and virtual staining workflow, illustrating the process from tissue sampling to pathology images. LPFM can achieve pathology image restoration and virtual staining in the workflow.     
    \textbf{b.} The unified architecture of LPFM integrates contrastive pre-training and prompt-guided conditional diffusion for task-specific generation.
    \textbf{c.} The curated datasets includes 87,810 whole-slide images (WSIs) and 190 million patches from 37 data sources, encompassing 34 tissue types and 5 staining protocols. 
    \textbf{d.} Our proposed LPFM was rigorously evaluated through 50 distinct experimental tasks organized into four fundamental image restoration categories, including 18 image super-resolution tasks (varying scale factors),  18 imaging deblurring tasks (different blurring kernel sizes), 18 Gaussian denoising tasks (with varying noise distributions), and 6 coupled-degradation pathology image restoration tasks (addressing composite artifacts).
    \textbf{e.} Performance evaluation of LPFM against competing methods across 6 virtual staining tasks, including Autofluorescence (AF) to Hematoxylin \& Eosin (H\&E) stains for rapid diagnosis, H\&E to Periodic Acid-Schiff-Alcian Blue (PAS-AB) for glycoprotein detection, H\&E to multiplex immunohistochemistry (mIHC) for biomarker analysis based on the original and degraded pathology images.
    }
    \label{fig: overview}
\end{figure*}

\section{Introduction}\label{sec1}
The advent of digital pathology has revolutionized modern medicine by transitioning traditional glass slides into high-resolution whole slide images (WSIs), enabling computerized analysis \cite{ma2025generalizable}, enhanced collaborative diagnostics across institutions \cite{chen2024towards}, and AI-assisted decision support \cite{song2023artificial}. This digital transformation began with slide scanning technologies \cite{bejnordi2017diagnostic} and has since evolved into an essential component of precision medicine \cite{ma2025pathbench,lai2023artificial}, allowing pathologists to examine tissue morphology at unprecedented scales \cite{xu2024multimodal} while facilitating large-scale collaborative research \cite{srinidhi2021deep,yan2025pathorchestra}. However, diagnostic utility is frequently compromised by multiple degradation problems in the imaging pipeline \cite{zhuang2025mim,liang2021swinir}. During slide preparation, tissue sections may suffer from folding \cite{jin2025hmil}, tearing \cite{echle2021deep}, or staining inhomogeneity \cite{wang2024rethinking}, while scanning introduces optical blur \cite{xiong2024takt}, noise \cite{zhang2017beyond}, and resolution limitations \cite{chen2024next}.

These degradations collectively obscure critical cellular features including nuclear pleomorphism \cite{bulten2020automated,tolkach2020high}, inflammatory infiltrates \cite{siemion2022we}, and subtle pathological changes \cite{coudray2018classification}, potentially leading to diagnostic uncertainty. The limitations of physical rescanning are manifold \cite{wang2024exploiting,yan2025pathorchestra}: time constraints \cite{bulten2020automated}, prohibitive costs \cite{krithiga2021breast}, technical irreversibility of preparation artifacts \cite{bejnordi2017diagnostic}, and frequent biopsy exhaustion \cite{xu2024multimodal}.
This has spurred computational approaches to enhance image quality through both restoration techniques (e.g., noise reduction \cite{zhuang2025mim}, deblurring \cite{xia2023diffir}, super-resolution \cite{ma2025pathbench}) and virtual staining methods \cite{guo2025focus,isola2017image}. Current solutions remain fragmented \cite{xie2024towards,chen2024next}, typically addressing either restoration or staining separately \cite{rombach2022high,qu2024tokenflow}, with denoising models potentially altering stain characteristics \cite{echle2021deep} and staining networks amplifying artifacts \cite{liang2023multi}.
The field lacks unified generative frameworks capable of handling diverse low-level vision tasks \cite{xiong2024autoregressive,li2025autoregressive}, forcing clinics to maintain incompatible specialized systems \cite{chen2024towards}. Pathology imaging particularly suffers from this fragmentation despite shared underlying challenges \cite{han2024infinity,fan2024fluid}.

Our work addresses this critical gap by introducing the first unified low-level generative foundation model for enhancing pathology image quality, including pathology image restoration and translation. Unlike existing task-specific models that process artifacts and stains in isolation, our approach recognizes that these challenges are fundamentally interconnected aspects of pathological image formation. This unified treatment enables synergistic improvements; for instance, stain-aware denoising and artifact-resistant stain transfer emerge naturally from the shared representation. We develop a novel pre-training paradigm using large-scale multi-tissue, multi-stain datasets, capturing both universal characteristics of pathological image degradation and stain-invariant morphological features. Furthermore, we pioneer a prompt-controlled framework that dynamically switches between low-level pathological tasks without architectural modifications. By integrating these innovations, we establish the first unified low-level pathology foundation model in computational pathology that moves beyond fragmented solutions to comprehensive image generation.

\section{Results}\label{results}
% 目前有了对应图像复原三个大类任务的3个图，还缺这样几个图
% 整体overview的图，差虚拟染色的几个图，因为总体的模式一致，所以这里使用一个图表述虚拟染色就可以了，这个图中最好从WSI级别的来分析，说明整体的虚拟染色能力，然后还需要一个说明prompt对整体模型调控能力的一张图，还需要一个对于下游任务的分析处理图，这里的下游任务应该是从high-level级别的任务出发来分析，说明复原的图像对于整体分类的能力有提升，这里可以从特征级别和结果级别统一画图说明。
% 1. Overview:
% b c d 用来说明现在用的数据集，其中要说明数据集的哪些要点，比如用于预训练的数据量多少，比如各种染色的情况是什么样的。
% b 说明现在数据集的预训练使用了哪些组织，哪些器官，包括哪些癌症，
% c 用于training， held-out， external的总体数据集包括多少
% d 说明染色类型包括哪些，不同类型的染色数量包括多少
% 2. 虚拟染色
% 三种虚拟染色任务的结果好坏，可视化效果是什么样的，图表分析如何，
% 3. Prompt的调控能力
% 使用不同prompt对整体效果的影响，使用不同prompt的结果是什么样的，可视化效果如何，效果对比情况如何
% 4. low-level对于下游任务的提升效果，特征和结果共同分析
% 对于下游High level任务的效果好坏，在存在退化情况下，使用low level模型过一遍的
% 加上3个图像复原任务的结果的图，一共是7个图在主体文章里面
In this section, we conducted comprehensive evaluations across 66 distinct experimental tasks, systematically organized into six fundamental low-level vision categories: (i) super-resolution (18 tasks with varying scale factors and degradation models), (ii) image deblurring (18 tasks covering different kernel sizes), (iii) image denoising (18 tasks with varying noise intensities), (iv) coupled degradation restoration (6 tasks addressing composite artifacts), (v) virtual staining (3 tasks for stain transformation), and (vi) virtual staining for degraded pathology image (3 tasks combining physical degradations with stain conversion).
Fig. \ref{fig: overview} presents a comprehensive overview of the proposed low-level pathology foundation model (LPFM) for pathology image restoration and virtual staining. As shown in Fig. \ref{fig: overview}b, LPFM included a contrastive pretraining framework that tackled multiple coupled degradation problems, generating coarsely restored and virtually stained images. The pretraining framework was trained through a contrastive loss by pulling closer the latent features of paired degraded images and their high-quality counterparts, while pushing away the features of unpaired samples.

Building on the coarsely restored images, we proposed a conditional diffusion model that improved image quality through a guided denoising process, utilizing both the coarse restorations and textual prompts as conditional inputs. In Fig. \ref{fig: overview}d-e, our LPFM demonstrated superior performance across all tasks, establishing itself as the first unified foundation model capable of handling multiple low-level vision challenges in computational pathology. The key advantages of LPFM include exceptional generalization across tissue types and staining protocols, and robust performance on isolated/coupled degradations (present tense for enduring conclusions).
To ensure a thorough assessment of image quality, we employed three complementary metrics: PSNR \cite{huynh2008scope} for pixel-level fidelity, SSIM \cite{wang2004image} for structural similarity, and LPIPS \cite{johnson2016perceptual} for perceptual quality. These metrics were consistently used in our experiments (Sec. \ref{Evaluation Metrics}) to provide quantitative comparisons. All experiments were conducted on partitioned datasets with 95\% confidence intervals and significance testing, ensuring reliable evaluation.

% Since all low-level tasks are ultimately aimed at obtaining better and more suitable pathology images for subsequent high-level tasks, we use the most classic WSI classification as our benchmark to compare the performance of various generative methods for downstream tasks. Specifically, we employ the same WSI classification method while using images generated by different tasks as input during the inference phase, and then compare their classification results to assess the effectiveness of each generative approach.

\begin{figure*}
    \centering
    \includegraphics[width=0.96\linewidth]{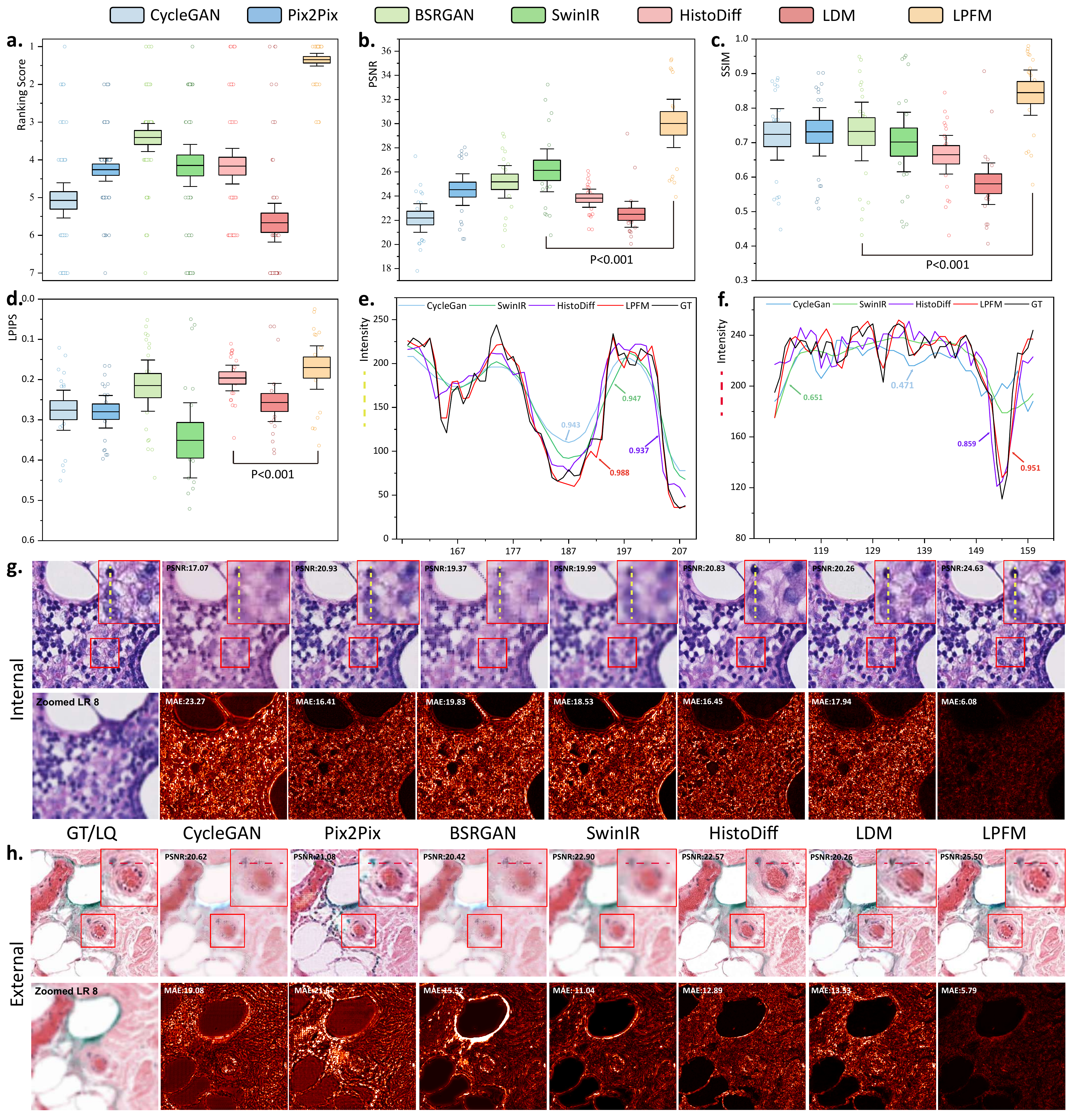}
    \caption{\textbf{Results of pathology image super resolution tasks.}  \textbf{a.} Average ranking of LPFM and compared methods based on PSNR, SSIM and LPIPS across 18 super resolution tasks.
    \textbf{b-d.} Average PSNR, SSIM and LPIPS of LPFM and compared methods across 18 super resolution tasks. Error bars represent 95\% CI. The box limits represent the standard error.
    \textbf{e-f.} Intensity profiles along the dashed yellow and red lines for the ground truth (GT) image and the top four performing models.
    The Pearson correlation coefficient (PCC) is provided for each method.
    \textbf{g-h.} The original GT images, 8 times downsampled images (LR 8) and restored images generated by various methods on internal and external datasets. Heatmaps of the mean absolute error (MAE) between the GT and generated images are also shown. Lower MAE indicates better performance.  
    }
    \label{fig: pipeline_sr_combined}
\end{figure*}

\subsection{Super Resolution}
\label{Super Resolution}
\begin{figure*}[t]
    \centering
    \includegraphics[width=\linewidth]{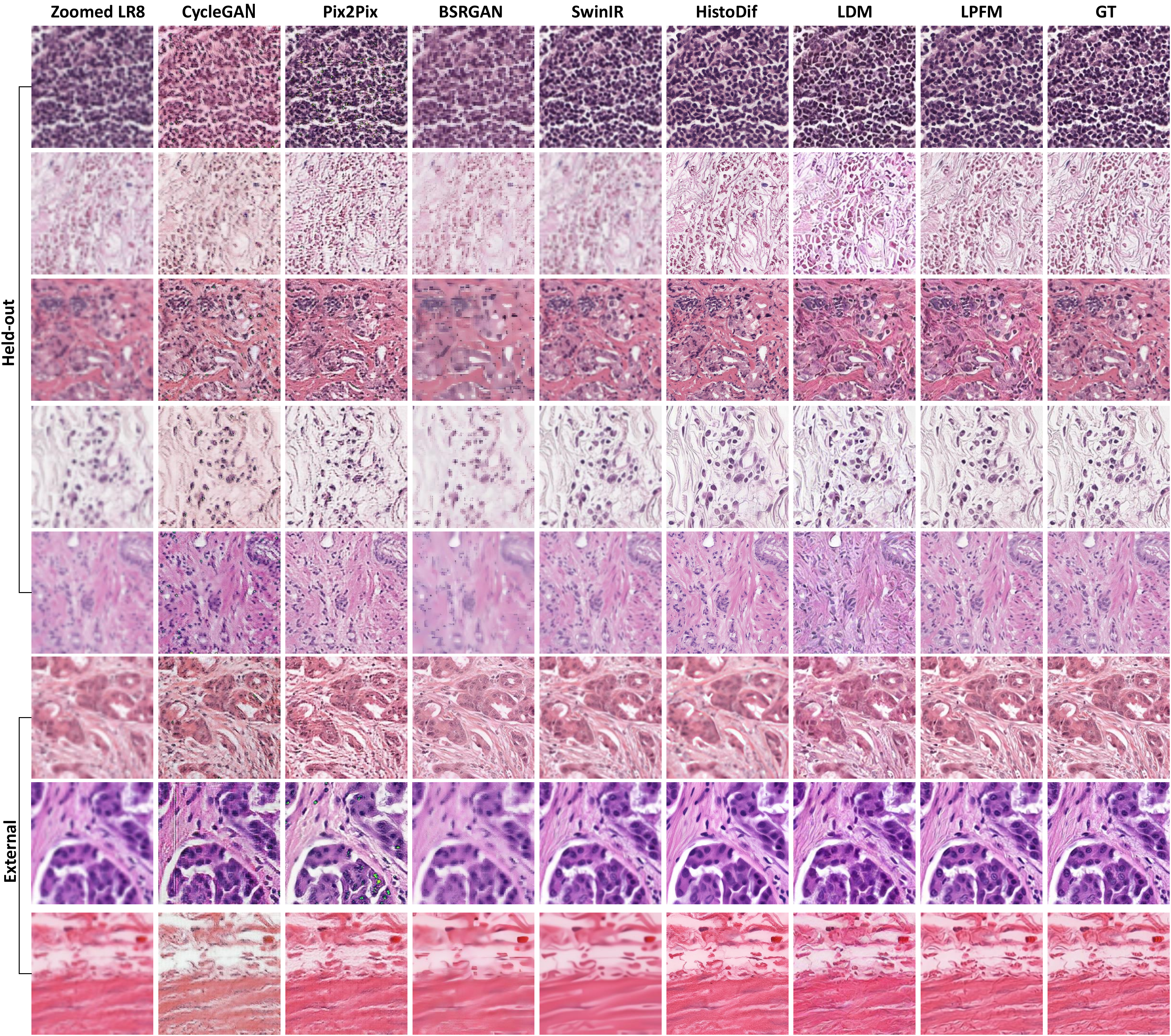}
    \caption{More samples of different methods for pathology image super resolution tasks. The original GT images, 8 times downsampled images (LR 8) and restored images generated by various methods on internal and external datasets. 
    }
    \label{fig: samplesSR}
\end{figure*}

% 首先一段说明super resolution的重要性，然后说明在这个任务中，我们使用的实验设置和数据集情况，在这些设置对应的实验任务中，我们取得的实验结果是什么样的，排名情况如何，相比于其他方法，我们方法在哪些领域中有不错的提升，提升多少，在多少个任务重取得了最好的效果，又因为哪些原因在某些任务中没有取得最好的效果。
Super resolution refers to computational techniques that reconstruct high-resolution images from low-resolution inputs by recovering lost high-frequency details and fine structures. In digital pathology, this process enhances the visibility of diagnostically critical features that may be obscured due to limitations in scanning resolution or image acquisition conditions. The ability to faithfully reconstruct these microscopic details is particularly important because pathologists routinely examine tissue specimens at multiple magnification levels, where fine cellular and subcellular features directly inform diagnostic decisions. Therefore, it is important to evaluate the super-resolution abilities of different models. In this section, we conducted experiments on a total of 18 tasks, including 9 internal tests and 9 external tests. The detailed experimental results are presented in Extended Data (Tab. \ref{tab:camelyon16_sr}-\ref{tab:ocelot_sr}). More generated samples are shown in Extended Data (Fig. \ref{fig: samplesSR}).

To simulate realistic image degradation for super-resolution evaluation, we generated low-resolution counterparts by downscaling original high-resolution pathology images by factors of 2$\times$, 4$\times$, and 8$\times$ using the comprehensive degradation model detailed in Sec. \ref{degradation_approach}. For internal validation, we employed three benchmark datasets (CAMELYON16\cite{litjens20181399}, PANDA\cite{bulten2022artificial}, and PAIP2020\cite{kim2023paip}), which were rigorously partitioned into training (70\%), validation (10\%), and test (20\%) sets with no data overlap to ensure unbiased evaluation. All model comparisons were performed exclusively on the held-out test sets under standardized conditions. To further validate generalization capability, we incorporated external test datasets (TIGER2021, MIDOG2022 and OCELOT\cite{Ryu_2023_CVPR}) representing diverse tissue types, staining protocols, and scanner variations, providing robust assessment across different clinical scenarios and imaging conditions.

To comprehensively evaluate the performance of different methods across multiple super-resolution tasks, we conducted a thorough ranking analysis where each of the 18 super-resolution tasks was ranked according to all three metrics (Fig. \ref{fig: pipeline_sr_combined}a). Our proposed method demonstrated superior and consistent performance across the comprehensive evaluation. In the 18 super-resolution tasks assessed by three metrics, our approach achieved an outstanding average ranking of 1.33 across all evaluation criteria (Fig. \ref{fig: pipeline_sr_combined}a). More impressively, in 15 out of these 18 tasks, our method simultaneously secured either first or second place rankings in all three metrics (PSNR, SSIM, and LPIPS). Specifically, the LPFM attained remarkable quantitative results with average values of 30.27 dB (PSNR), 0.85 (SSIM), and 0.1647 (LPIPS) across all tasks, surpassing the second-best methods by significant margins of 4.14 dB and 0.12 in PSNR and SSIM, respectively (Fig. \ref{fig: pipeline_sr_combined}b-d).

Furthermore, we showed some generated visual samples of various methods in internal and external datasets (Fig. \ref{fig: pipeline_sr_combined}g-h). The mean absolute error (MAE) of the generated images and GT images was computed to evaluate the visual quality of the generated images. LPFM achieved the lowest MAE for the samples in internal and external samples. Additionally, we analyzed the intensity profiles of the generated samples in internal and external datasets produced by the top four best-performance models alongside the GT images (Fig. \ref{fig: pipeline_sr_combined}e-f). To better validate the fidelity of reconstructed details, we conducted pixel-wise Pearson Correlation analysis \cite{benesty2009pearson} between the generated images and GT images. The results demonstrated that LPFM achieved the highest correlation coefficient. As shown in Fig. \ref{fig: pipeline_sr_combined}e-f, the intensity curve of LPFM-generated images exhibited nearly perfect alignment with the GT profile, particularly in preserving critical high-frequency components that correspond to cellular boundaries and nuclear details. The quantitative correlation analysis, combined with our previous metric evaluations, provided comprehensive evidence that LPFM delivered both perceptually convincing and accurate super-resolution results for pathology images.

\subsection{Image Deblurring}

\begin{figure*}
    \centering
    \includegraphics[width=0.98\linewidth]{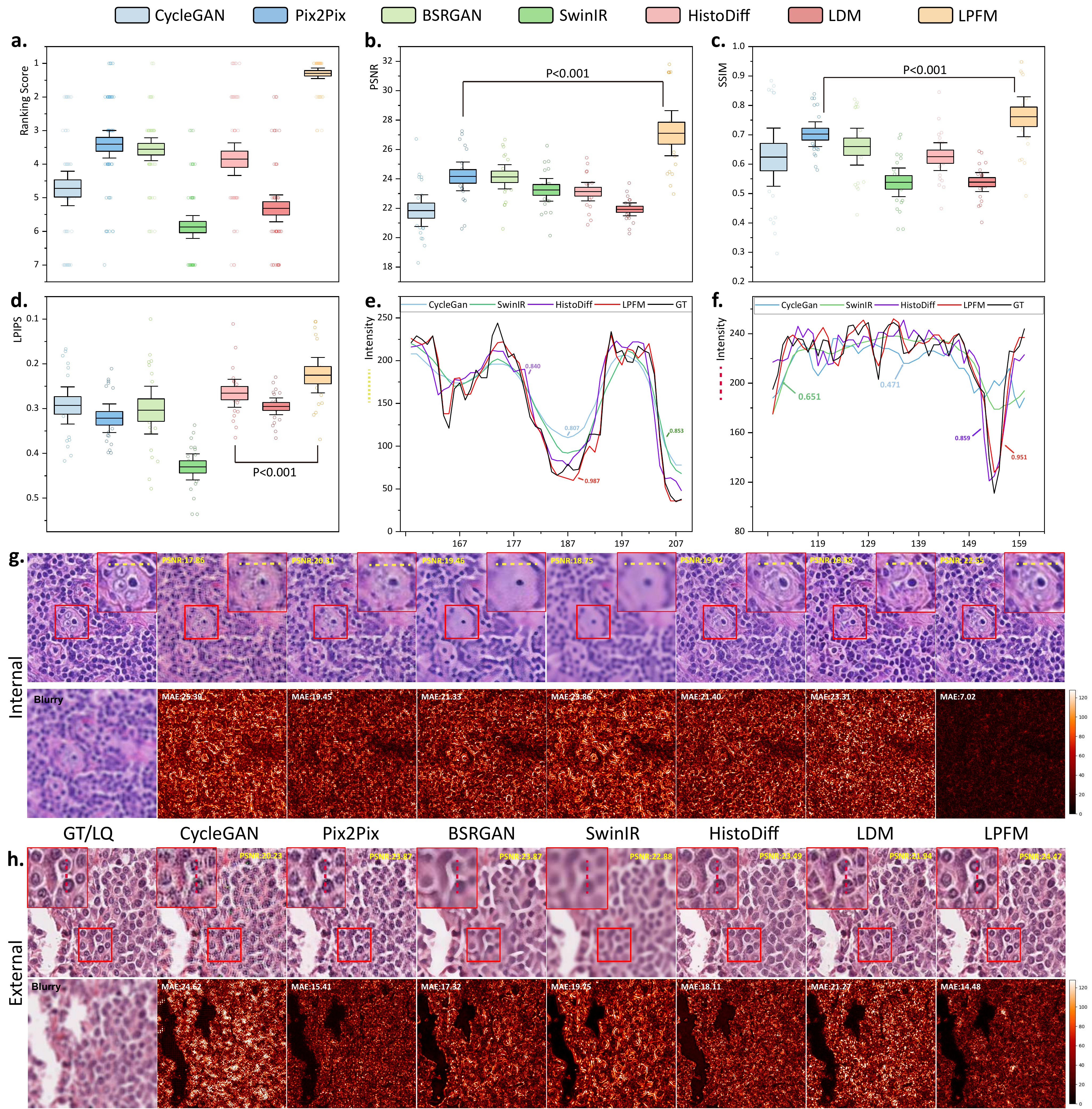}
    \caption{\textbf{Results of pathology image deblurring tasks.}  
    \textbf{a.} Average ranking of LPFM and compared methods based on PSNR, SSIM and LPIPS across 18 deblurring tasks.
    \textbf{b-d.} Average PSNR, SSIM and LPIPS of LPFM and compared methods across 18 deblurring tasks. Error bars represent 95\% CI. The box limits represent the standard error.
    \textbf{e-f.} Intensity profiles along the dashed yellow and red lines for the GT image and the top four performing models. PCC is provided for each method.
    \textbf{g-h.} The original GT images, blurry images with 15 pixel Gaussian kernel and restored images generated by various methods on internal and external datasets. A heatmap of the MAE between the GT and generated images is also shown.
    }
    \label{fig: pipeline_db_combined}
\end{figure*}
In computational pathology, effective deblurring enhances critical diagnostic features by restoring sharp boundaries and fine cellular details that may be lost due to optical limitations or focus variations during slide scanning. This capability directly impacts diagnostic accuracy in applications such as tumor margin assessment and mitotic figure detection.

\begin{figure*}[t]
    \centering
    \includegraphics[width=\linewidth]{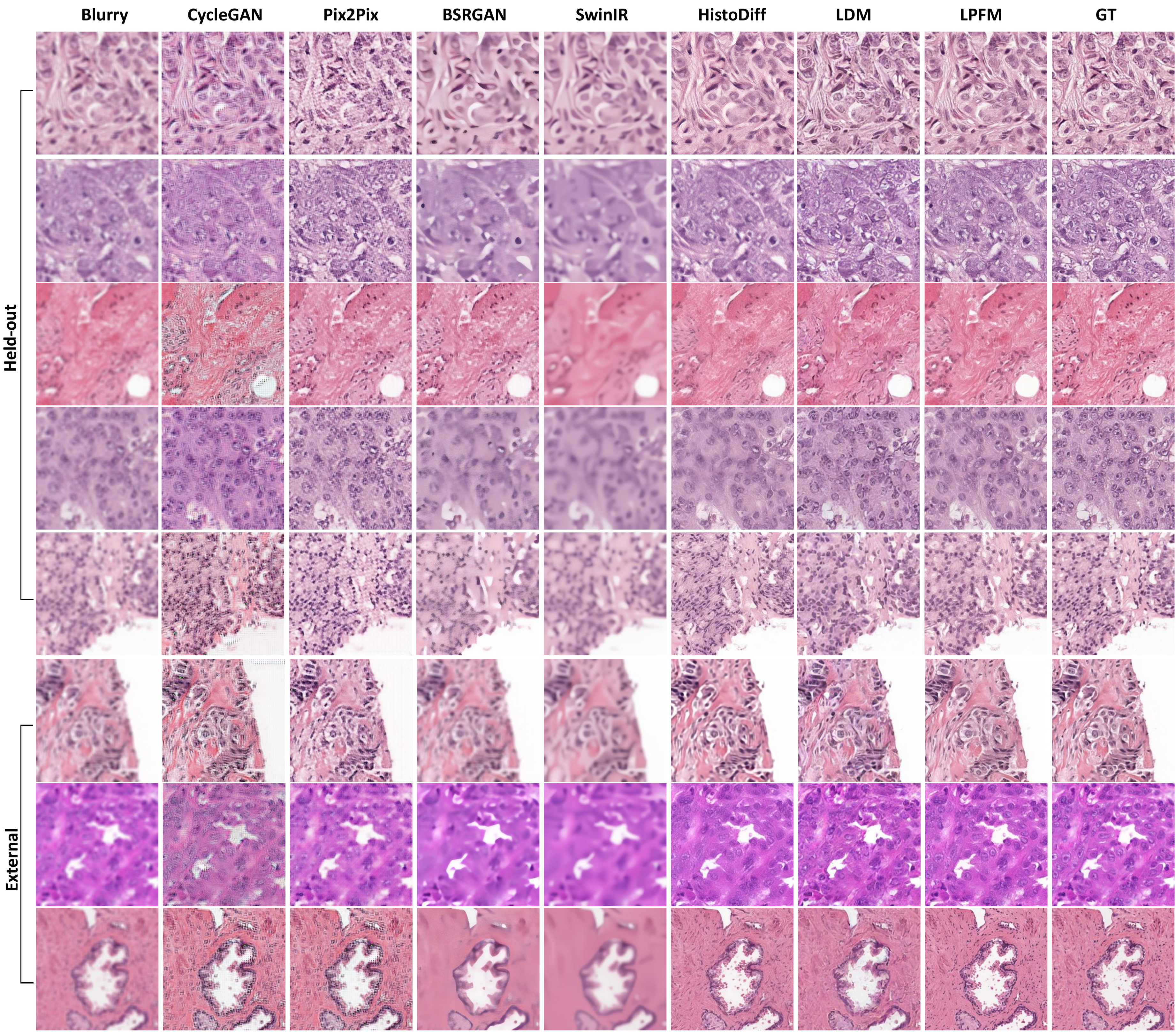}
    \caption{More samples of different methods for pathology image deblurring tasks. The original GT images, blurry images with 15 pixel Gaussian kernel (Blur 15) and restored images generated by various methods on internal and external datasets.
    }
    \label{fig: samplesDB}
\end{figure*}

Our evaluation framework comprised 18 deblurring tasks (9 internal and 9 external tests), with complete results available in Extended Data (Tab. \ref{tab:camelyon16_noise}-\ref{tab:ocelot_noise}). We simulated clinically relevant blur conditions using Gaussian kernels with varying parameters (kernel sizes: 7-15, $\sigma_1, \sigma_2$: 1.5-3.5) as detailed in Sec. \ref{degradation_approach}. The same rigorous dataset partitioning (CAMELYON16, PANDA, PAIP2020) and external validation protocol (OCELOT, MIDOG2022, TIGER2021) described in Sec. \ref{Super Resolution} were maintained. More generated samples are shown in Extended Data (Fig. \ref{fig: samplesDB}).

The statistical analysis showed that LPFM achieved the best average ranking scores across PSNR, SSIM and LPIPS metrics for the 18 deblurring tasks (Fig. \ref{fig: pipeline_db_combined}a). Specifically, LPFM ranked among the top two methods in all three metrics for 16 out of 18 tasks (84.2\% of cases), demonstrating remarkable consistency across different evaluation criteria. Notably, LPFM achieved the highest PSNR values in all 18 tasks, with an average score of 27.36 dB that significantly outperformed the second-best method (24.17 dB, +3.19 dB improvement). For structural similarity assessment, LPFM maintained superior performance with SSIM scores consistently leading all comparison methods across various blur conditions (Fig. \ref{fig: pipeline_db_combined}c). The average SSIM of 0.770 substantially exceeded competing approaches, with particularly notable advantages in challenging cases involving large kernel sizes where LPFM achieved up to 0.948 SSIM versus 0.824-0.845 for other methods. Perceptual quality evaluation through LPIPS further confirmed LPFM's advantages (Fig. \ref{fig: pipeline_db_combined}d), with an average score of 0.220 representing 31.5\% reduction in perceptual distance compared to traditional methods (CycleGAN: 0.293) and 17.3\% improvement over recent diffusion-based approaches (HistoDiff: 0.266).

Visual comparisons presented in Fig. \ref{fig: pipeline_db_combined}g-h showed LPFM consistently producing sharper cellular boundaries and better-preserved nuclear details compared to other methods. Quantitative analysis revealed LPFM achieved the lowest MAE across internal and external datasets, with internal samples showing 7.02 (versus 19.45 for the second-best method) and external datasets demonstrating 14.48 (versus 15.41 for the second-best method). The evaluation of structural consistency through the pixel-level PCC analysis further confirmed LPFM's exceptional performance, achieving 0.987 on internal datasets and maintaining strong generalization with 0.953 on external datasets (Fig. \ref{fig: pipeline_db_combined}e-f). These results collectively validated LPFM's capability to faithfully restore diagnostically critical features while maintaining structural consistency.

\subsection{Image Denoising}
\begin{figure*}
    \centering
    \includegraphics[width=0.98\linewidth]{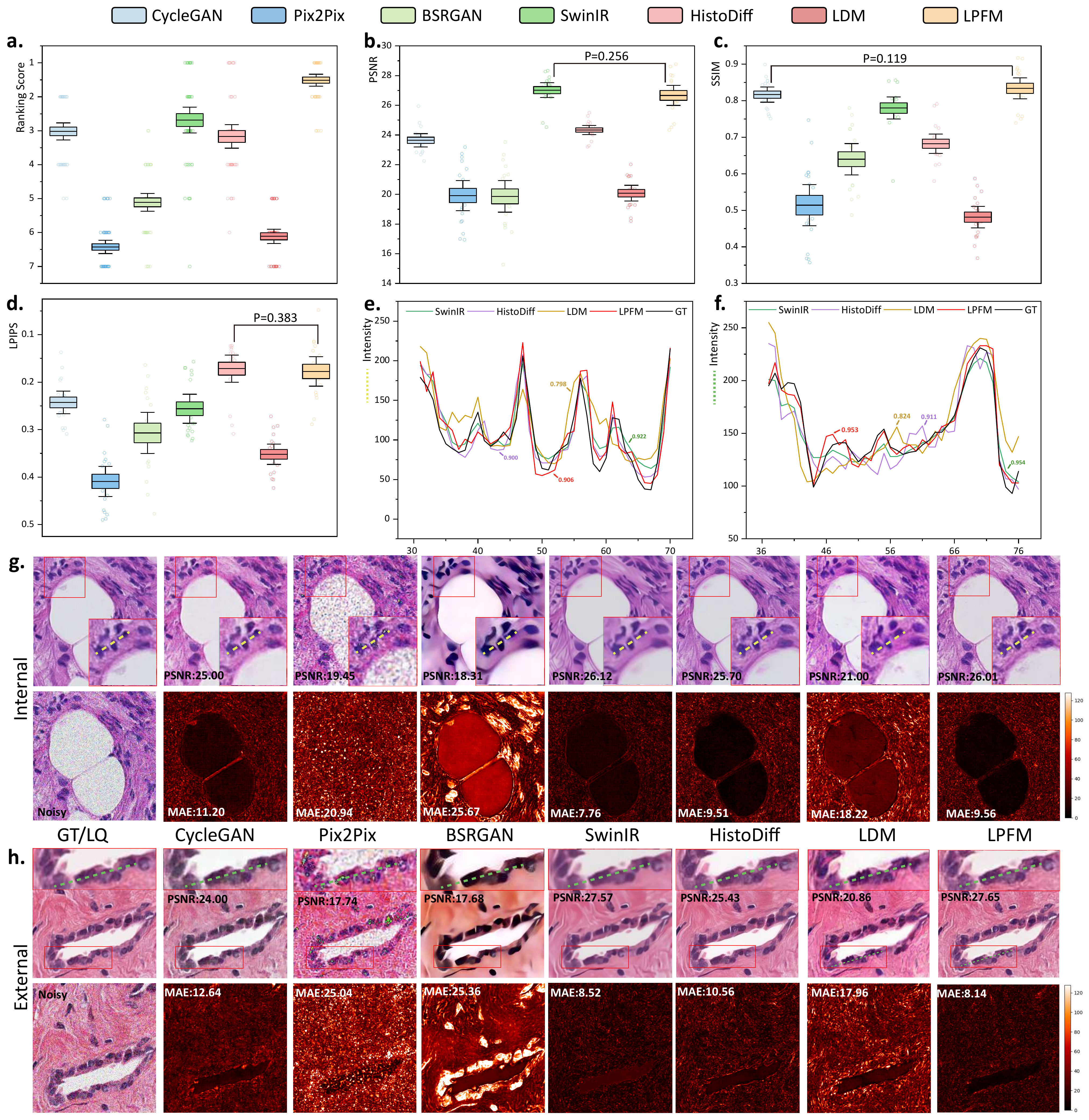}
    \caption{\textbf{Results of pathology image denoising tasks.  }
    \textbf{a.} Average ranking of LPFM and compared methods based on PSNR, SSIM and LPIPS across 18 denoising tasks.
    \textbf{b-d.} Average PSNR, SSIM and LPIPS of LPFM and compared methods across 18 denoising tasks. Error bars represent 95\% CI. The box limits represent the standard error.
    \textbf{e-f.} Intensity profiles along the dashed yellow and green lines for the GT image and the top four performing models. PCC is provided for each method.
    \textbf{g-h.} The original GT images, noisy images with additive Gaussian noise ($\sigma$=41) and restored images generated by various methods on internal and external datasets. A heatmap of the MAE between the GT and generated images is also shown.
    }
    \label{fig: pipeline_dn_combined}
\end{figure*}
\begin{figure*}[t]
    \centering
    \includegraphics[width=\linewidth]{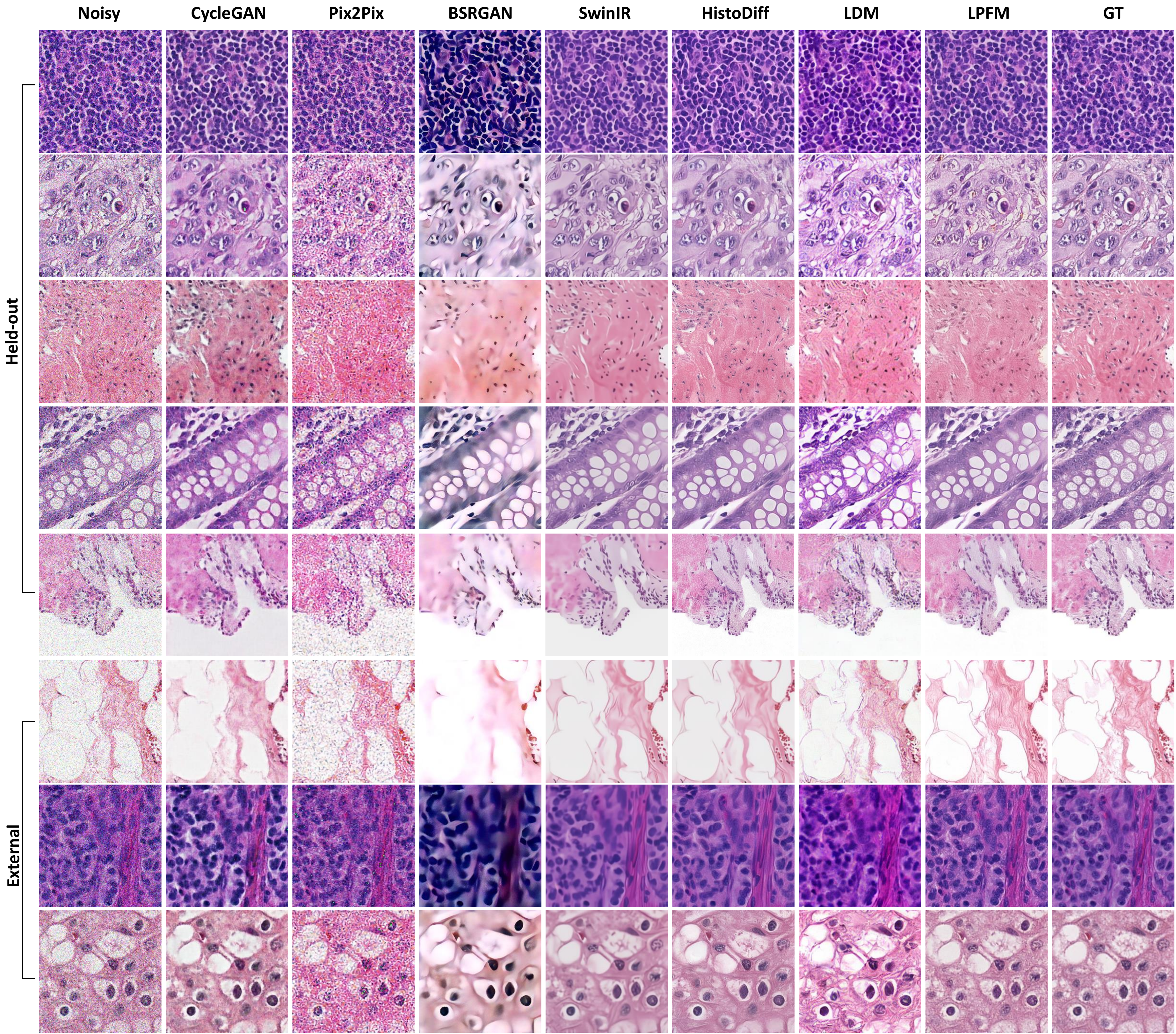}
    \caption{More samples of different methods for pathology image denoising tasks. The original GT images, noisy images with additive Gaussian noise ($\sigma$=41) (Noise 41) and restored images generated by various methods on internal and external datasets.
    }
    \label{fig: samplesDN}
\end{figure*}

Noise corruption in pathology images presents a significant challenge for both clinical diagnosis and computational analysis. Multiple factors introduce noise, including electronic sensor limitations during slide scanning, uneven staining artifacts, tissue preparation inconsistencies, and optical imperfections in imaging systems. These noise patterns obscure critical cellular and subcellular features essential for accurate diagnosis, including nuclear membrane integrity, chromatin distribution patterns, and subtle morphological characteristics. Therefore, it is necessary to remove possible noise inside pathology images and restore high-quality images for downstream tasks. To evaluate the noise restoration performance of various methods, we conducted 18 denoising experiments, including 9 internal and 9 external tests. The detailed experimental results are presented in Extended Data (Tab. \ref{tab:camelyon16_deblur}-\ref{tab:ocelot_deblur}).  More generated samples are shown in Extended Data (Fig. \ref{fig: samplesDN}).

To evaluate denoising performance, we generated synthetic datasets by corrupting high-quality pathology images with Gaussian noise at varying intensities ($\sigma$ = 21, 31, 41), following the degradation model in Sec. \ref{degradation_approach}. We used the same internal datasets (CAMELYON16, PANDA, PAIP2020) with a 7:1:2 train/val/test split and further validated generalization on external datasets (OCELOT, MIDOG2022, TIGER2021).

The ranking analysis of 18 denoising tasks revealed the superior performance of LPFM (Fig. \ref{fig: pipeline_dn_combined}a), which achieved an outstanding average ranking score of 1.48, significantly outperforming the second-best method SwinIR (average ranking 2.65) by 1.17. LPFM ranked among the top two methods in all three evaluation metrics (PSNR, SSIM, and LPIPS) for 14 out of the 18 tasks, showcasing its remarkable robustness in balancing different aspects of image quality. The substantial performance gap between LPFM and competing methods was particularly evident in high-noise scenarios ($\sigma$=41) where LPFM maintained superior detail preservation while effectively suppressing noise artifacts.

The comprehensive evaluation across all 18 denoising tasks revealed an interesting performance landscape among the compared methods (Fig. \ref{fig: pipeline_dn_combined}b-d). While LPFM did not achieve the highest scores in PSNR (SwinIR led with 27.02 dB average) or LPIPS (HistoDiff led with 0.172 average), it demonstrated exceptional balance across all three metrics (PSNR, SSIM, and LPIPS). Crucially, LPFM's superior SSIM (average 0.837), which measures structural similarity, indicated that it best maintained critical tissue and cellular details essential for pathological diagnosis, even if its PSNR or LPIPS was marginally lower than the best performing methods. This balance is vital in medical imaging where over-smoothing (high PSNR but loss of detail) or perceptual artifacts (good LPIPS but unnatural textures) can compromise diagnostic accuracy.

Visual and quantitative analysis of denoising performance was provided in Fig. \ref{fig: pipeline_dn_combined}g-h. In terms of MAE metric, LPFM showed differentiated performance across internal and external datasets. On internal test sets, LPFM achieved an MAE of 9.56, slightly higher than SwinIR's 7.76. However, in external validation datasets, LPFM showed superior generalization with the lowest MAE of 8.14, outperforming SwinIR's 8.52. Structural consistency analysis through the pixel-level PCC (Pearson Correlation Coefficient) showed similar trends. On internal data, LPFM achieved a high PCC of 0.906 (second to SwinIR's 0.922), while on external data it reached 0.953 - nearly identical to SwinIR's 0.954. These results indicated that both methods preserved local structures well, with SwinIR having a marginal advantage on familiar data distributions. While SwinIR achieved strong quantitative scores, its outputs frequently exhibited over-smoothing artifacts that erased diagnostically important cellular details and tissue textures. In contrast, LPFM maintained superior perceptual quality, preserving nuclear boundaries and chromatin patterns, even if some pixel-level metrics showed slight disadvantages.

\subsection{Virtual Staining}
\begin{figure*}
    \centering
    \includegraphics[width=0.98\linewidth]{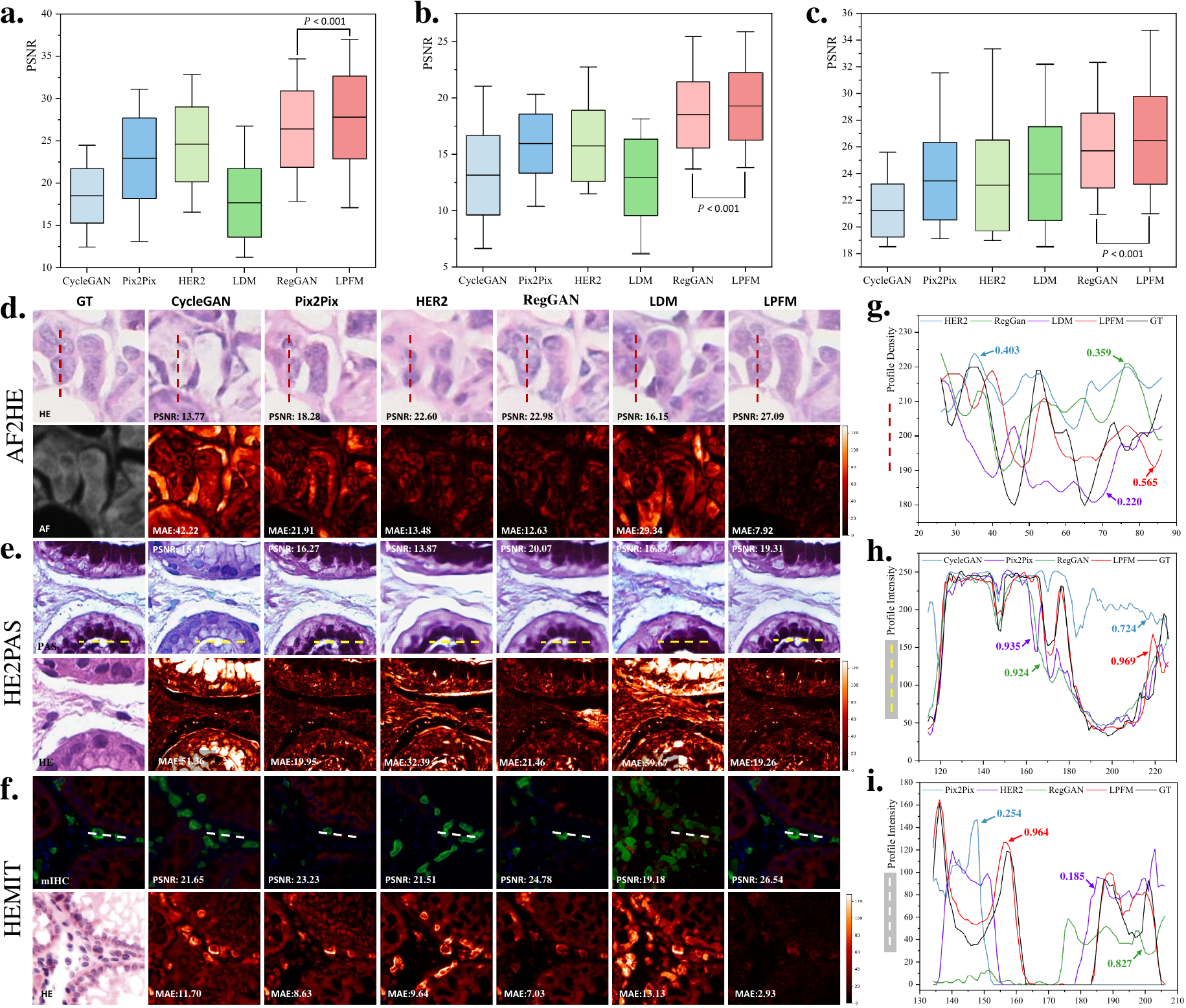}
    \caption{\textbf{Results of pathology image virtual staining tasks.  }
    \textbf{a-c.} Average PSNR of LPFM and compared methods based on AF2HE, HE2PAS and HEMIT datasets.  If LPFM outperforms the second-best method, the p-value is also reported. The box limits represent the standard deviation, and the error bars indicate the 2.5\% and 97.5\% percentiles.
    \textbf{d.} The original AF image, chemically H\&E-stained image, and virtually stained images generated by various methods on AF2HE dataset.
    \textbf{e.} The H\&E-stained image, chemically stained PAS-AB image, and virtually stained images generated by various methods on HE2PAS dataset.
    \textbf{f.} The H\&E-stained, mIHC, and corresponding virtually stained images predicted by various methods on HEMIT dataset. The second rows display the mean absolute error (MAE) heatmap between the chemical stained images and the virtually stained images.
    \textbf{g-i.} Intensity profiles along the dashed red, yellow and white lines for the ground truth image and the top five performing models. The Pearson correlation coefficient (PCC) is provided for each method.
    }
    \label{fig: pipeline_vs_combined}
\end{figure*}

% 这里首先要介绍一下这几个数据集是怎么样的，是怎么染色的，这几种染色的优势和好处是什么，为什么要选择这几个数据集做虚拟染色的比对？然后说明虚拟染色的好处和优势，再分别介绍一下这几个数据集的特点和数据量什么的，我们是怎么处理的。这里的对比方法也有所不同，这几个对比方法是什么。接着对图中的结果进行阐述说明，说明我们方法的优势。这一段有点长，如果有必要分小点说明吧。
Virtual staining, enabled by AI models, offers a transformative approach in pathology by digitally replicating the appearance of chemically stained tissue samples without the need for physical dyes. This technology significantly accelerates diagnostic workflows, generating high-quality stained images in minutes rather than the hours or days required for traditional chemical staining methods, while also reducing costs associated with reagents and laboratory labor.

\begin{figure*}[t]
    \centering
    \includegraphics[width=\linewidth]{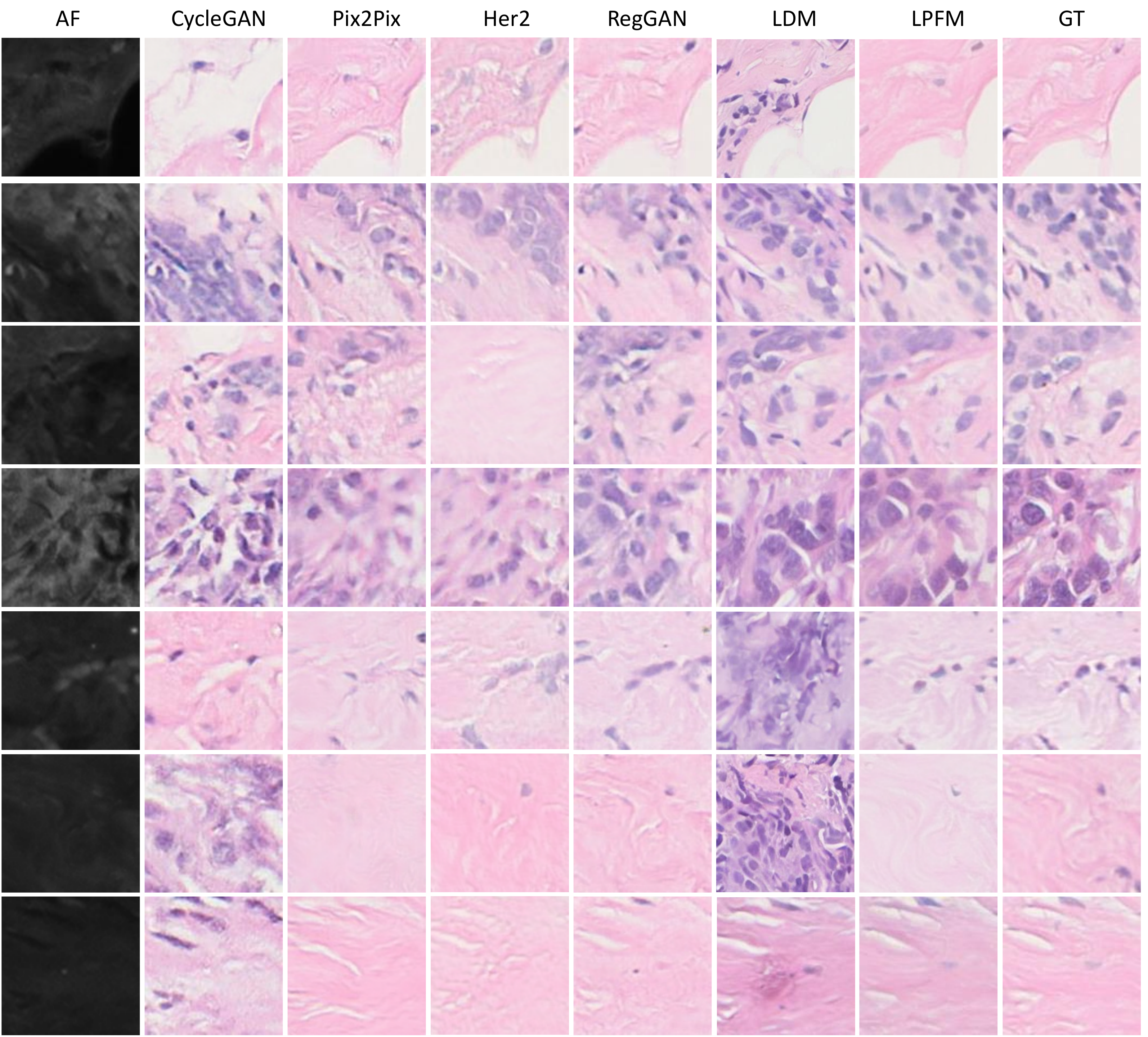}
    \caption{More samples of different methods for virtual staining on AF2HE dataset (autofluorescence to H\&E stain). The autofluorescence images, GT (H\&E) images and virtually stained images generated by various methods are presented.
    }
    \label{fig: samples_af2he}
\end{figure*}

\begin{figure*}[t]
    \centering
    \includegraphics[width=\linewidth]{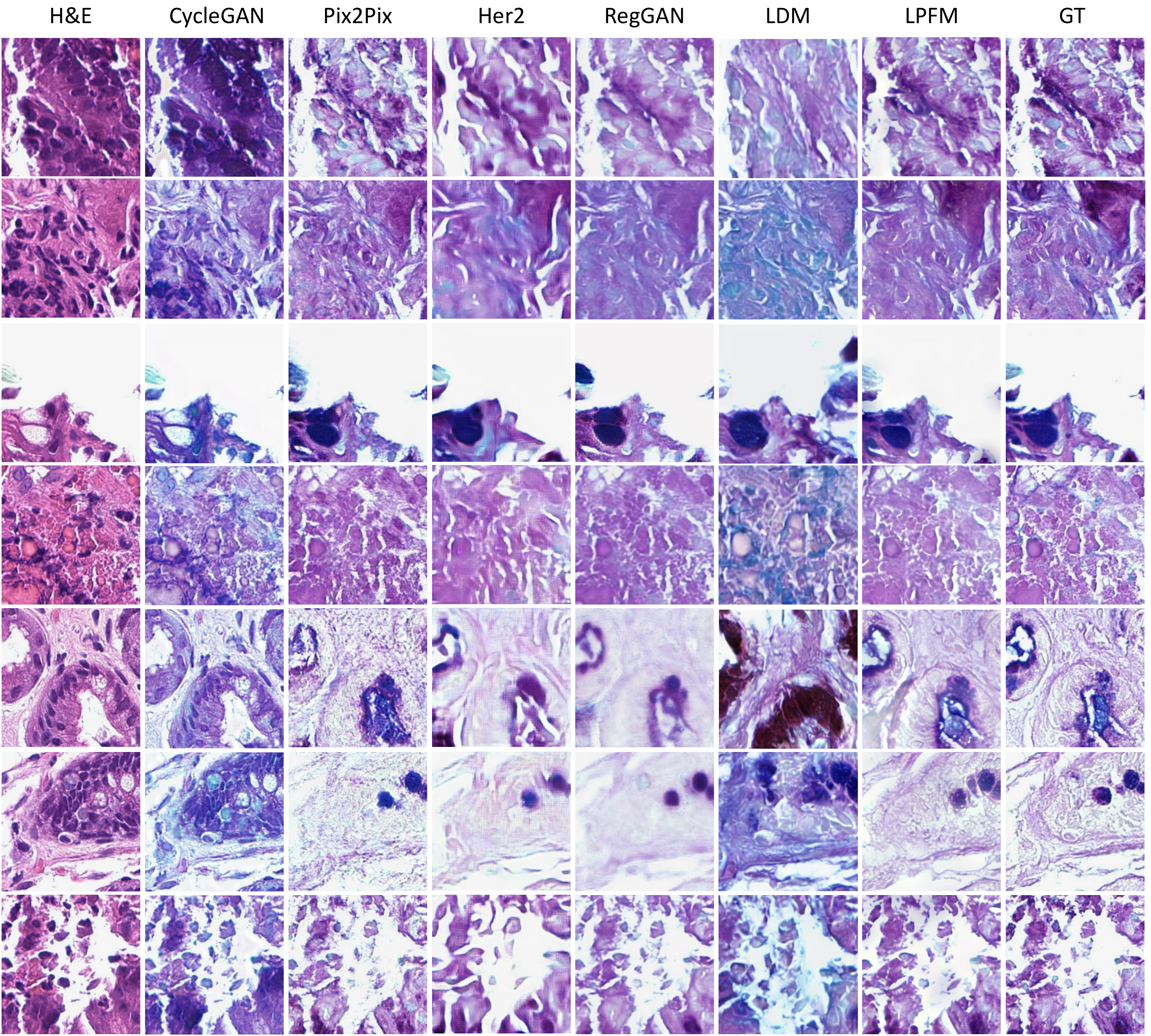}
    \caption{More samples of different methods for virtual staining on HE2PAS dataset, including the paired H\&E stained and Periodic Acid-Schiff-Alcian Blue (PAS-AB) stained images. The H\&E images, GT (PAS-AB) images and virtually stained images generated by various methods are presented.
    }
    \label{fig: samples_he2pas}
\end{figure*}

\begin{figure*}[t]
    \centering
    \includegraphics[width=\linewidth]{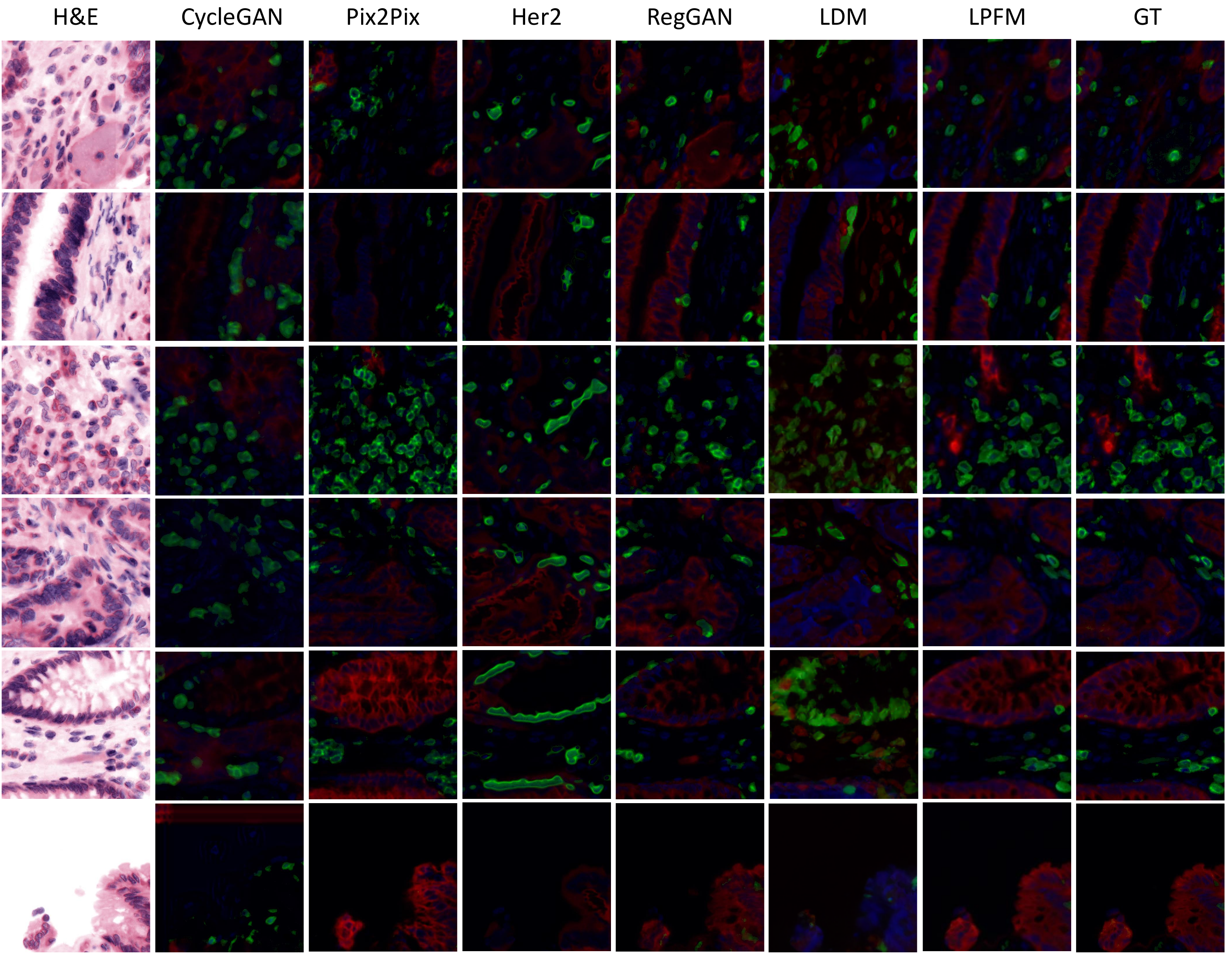}
    \caption{More samples of different methods for virtual staining on HEMIT dataset, including the paired H\&E stained and multiplex immunohistochemistry (mIHC) stained images. The H\&E images, GT (mIHC) images and virtually stained images generated by various methods are presented.
    }
    \label{fig: samples_hemit}
\end{figure*}

For the virtual staining tasks, we employed multiple paired staining datasets, including AF2HE \cite{dai2022weakly}, HE2PAS, and HEMIT \cite{bian2021immunoaizer} datasets, to rigorously validate the performance of LPFM and various compared methods. Each dataset served a distinct purpose: AF2HE evaluated the model's ability to transform autofluorescence (AF) images into H\&E stains, crucial for rapid preliminary diagnostics; HE2PAS assessed the conversion between H\&E and Periodic Acid-Schiff-Alcian Blue (PAS-AB) stains, important for detecting glycoproteins and mucins in conditions like kidney and liver diseases; and HEMIT tested the model's capability to predict multiplex immunohistochemistry (mIHC) staining from H\&E, enabling advanced biomarker analysis without repeated physical staining. These datasets were selected to cover diverse staining modalities and clinical scenarios, ensuring robust validation across different tissue structures and diagnostic needs. The detailed experimental results are presented in Extended Data (Tab. \ref{tab:af2he_vs}-\ref{tab:hemit_vs}).
 More generated samples are shown in Extended Data (Fig. \ref{fig: samples_af2he}-\ref{fig: samples_hemit}).

To comprehensively evaluate our approach, we compared LPFM against several widely used methods for virtual staining tasks, each with distinct architectures and advantages. CycleGAN, an unsupervised generative adversarial network (GAN), excelled in unpaired image-to-image translation through its cyclic consistency loss, making it suitable when strictly paired training data was unavailable. Pix2Pix, a conditional GAN, leveraged paired data for precise pixel-to-pixel translation, offering superior performance in scenarios where exact input-output alignments were critical. HER2 specialized in histopathology image translation by incorporating hierarchical feature extraction, enhancing structural preservation in complex tissue architectures. RegGAN introduced a registration-based loss to improve spatial alignment between input and output images, particularly beneficial for maintaining morphological accuracy in virtual staining tasks. Lastly, Latent Diffusion Models (LDM) employed a denoising diffusion process in latent space, combining the generative power of diffusion models with computational efficiency.

Our proposed LPFM demonstrated superior performance across all three virtual staining datasets (AF2HE, HE2PAS, and HEMIT) when compared to existing methods, as evidenced by both quantitative metrics and qualitative assessments. Quantitatively, LPFM achieved the highest average PSNR values (Fig. \ref{fig: pipeline_vs_combined}a-c). Specifically, in the AF2HE task, LPFM achieved a PSNR of 27.81 dB, representing a 5.3\% improvement over the second-best method RegGAN (26.42 dB). The superior performance was further confirmed in the HE2PAS and HEMIT tasks where LPFM attained PSNR values of 19.29 dB and 26.51 dB respectively, corresponding to 4.1\% and 3.0\% improvements over RegGAN.

The architectural innovations in LPFM yielded substantial benefits in both structural preservation and perceptual quality. For structural similarity, LPFM achieved SSIM scores of 0.763, 0.563, and 0.820 across the three tasks, outperforming the second-best methods by 4.5\%, 33.8\%, and 10.7\% respectively. The perceptual quality metrics (LPIPS) showed consistent advantages, with LPFM demonstrating 20.1\%, 7.0\%, and 20.0\% reductions in perceptual error compared to the leading alternatives for each task.

Qualitatively, LPFM-generated virtual stains (Fig. \ref{fig: pipeline_vs_combined}d-f) exhibited remarkable fidelity to chemical staining results, with significantly reduced artifacts and better preservation of critical diagnostic features compared to other methods. The MAE heatmaps revealed that LPFM produced the smallest errors in challenging regions such as cell nuclei boundaries (H\&E), glomerular basement membranes (PAS-AB), and biomarker expression patterns (mIHC). The intensity profile analyses (Fig. \ref{fig: pipeline_vs_combined}g-i) further confirmed this advantage, with LPFM showing the highest Pearson correlation coefficients (PCC) with ground truth stained images, particularly in capturing fine-grained histological features.

\begin{figure*}
    \centering
    \includegraphics[width=0.98\linewidth]{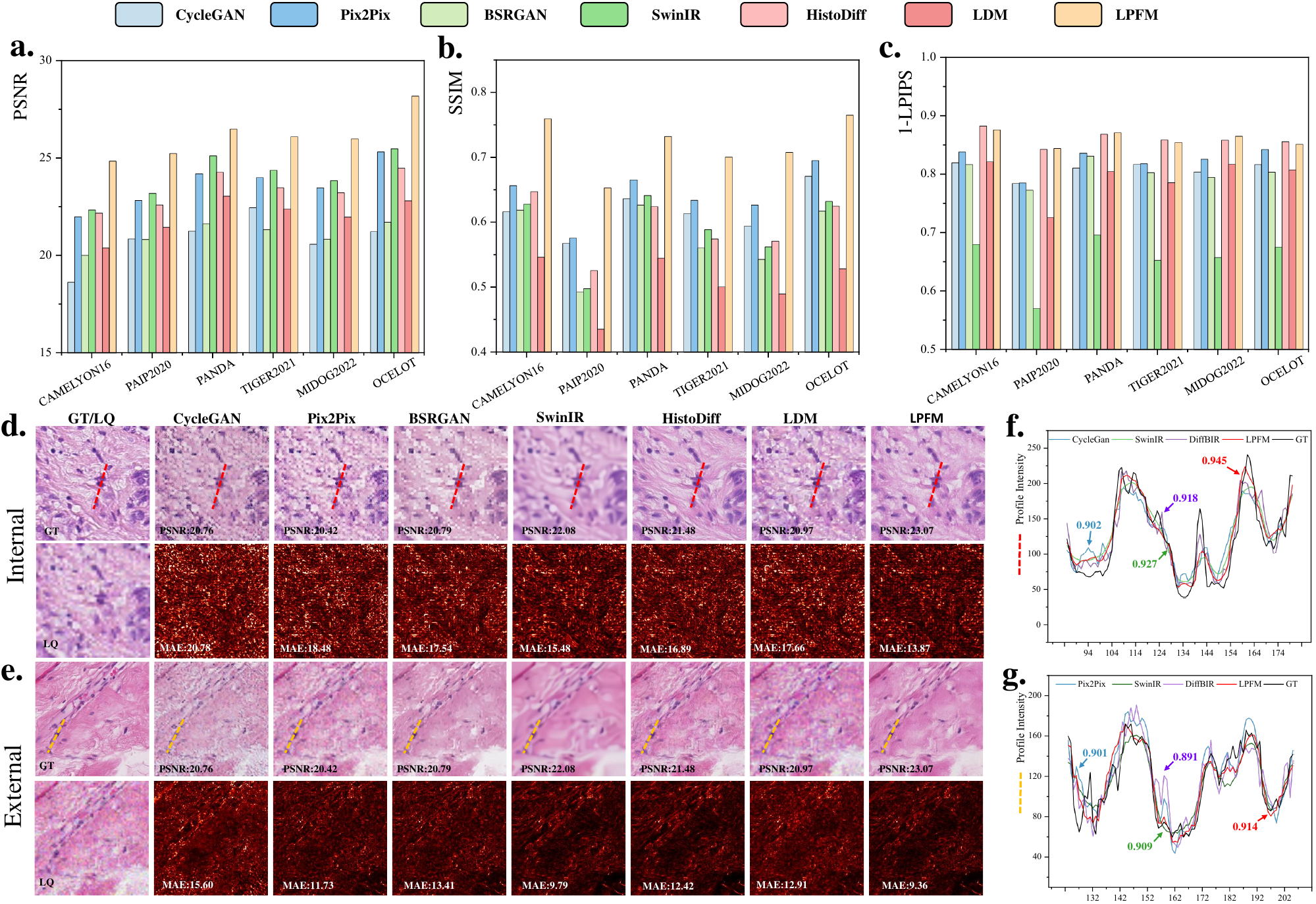}
    \caption{\textbf{Restoration results of pathology images with coupled degradations.}
    \textbf{a-c.} Average PSNR, SSIM and LPIPS of LPFM and compared methods on CAMELYON16, PAIP2020, PANDA, MIDOG2022, TIGER2021 and OCELOT datasets.
    \textbf{d-e.} The high-quality ground truth pathology images, degraded low-quality pathology images, mean average error images and restored pathology images generated by various methods on internal CAMELYON16 and external TIGER2021 datasets.
    \textbf{g-h.} Intensity profiles along the dashed red and orange lines for the GT image and the top four performing models. PCC is provided for each method.
    }
    \label{fig: coupled_degraded_results}
\end{figure*}
\subsection{Restoration for Coupled Degradations}
% Coupled LowLevel tasks
% 怎么展示这里。我首先预估这里应该是有两种不同的退化耦合。一种是不同退化类型的耦合，一种是结合虚拟染色和退化的耦合。前者比较好处理，直接使用RealESR那个代码，生成不同退化的耦合结果就好，其中使用随机化处理，生成一个在Internal（Held out）数据集上的测试集，一个在Extrernal上的测试集，直接在这两个测试集上做实验，看恢复效果。后者则是原染色上做某一个类型的退化吧，再做虚拟染色和复原，看结果怎么样，就做这两个就差不多了。

% In clinical practice, pathology images often exhibit coupled degradations arising from the interplay of multiple distortion types, such as blur, noise, low resolution, etc. While existing methods have demonstrated good performance in addressing isolated degradation types, their effectiveness diminishes significantly when applied to composite degradations. This limitation stems from two primary factors: task interferencewhere optimizing for one degradation type may exacerbate others, and distribution mismatchwhere models trained on individual degradations fail to capture the complex nonlinear interactions present in real-world scenarios.

% To rigorously evaluate model robustness under such challenging conditions, we constructed a comprehensive degradation framework by applying multiple sequential distortions to high-quality pathology images from datasets. This framework addresses two principal degradation scenarios. The first involves type-coupled degradations through randomized combinations of Gaussian blur, Poisson noise and low resolution with random parameters sampled from clinical imaging distributions. The second examines virtual staining-coupled degradations where originally stained images first undergo targeted quality reduction before virtual staining and restoration processes.
Real-world pathology images often suffer from multiple coexisting degradations, including blur, noise, and low resolution. While existing methods perform well on single degradation types, their effectiveness significantly decreases when handling such composite cases. This performance drop primarily occurs because optimizing for one degradation type may interfere with addressing others, and models trained on isolated degradations fail to capture the complex interactions present in actual clinical images.

To rigorously evaluate model robustness under such challenging conditions, we constructed a comprehensive degradation framework by applying multiple sequential distortions to high-quality pathology images from our datasets. These distortions included randomized combinations of Gaussian blur, Poisson noise, and low resolution with parameters carefully sampled to reflect real clinical imaging conditions. The degradation process preserved the biological relevance of the images while introducing realistic artifacts.

\begin{figure*}[t]
    \centering
    \includegraphics[width=\linewidth]{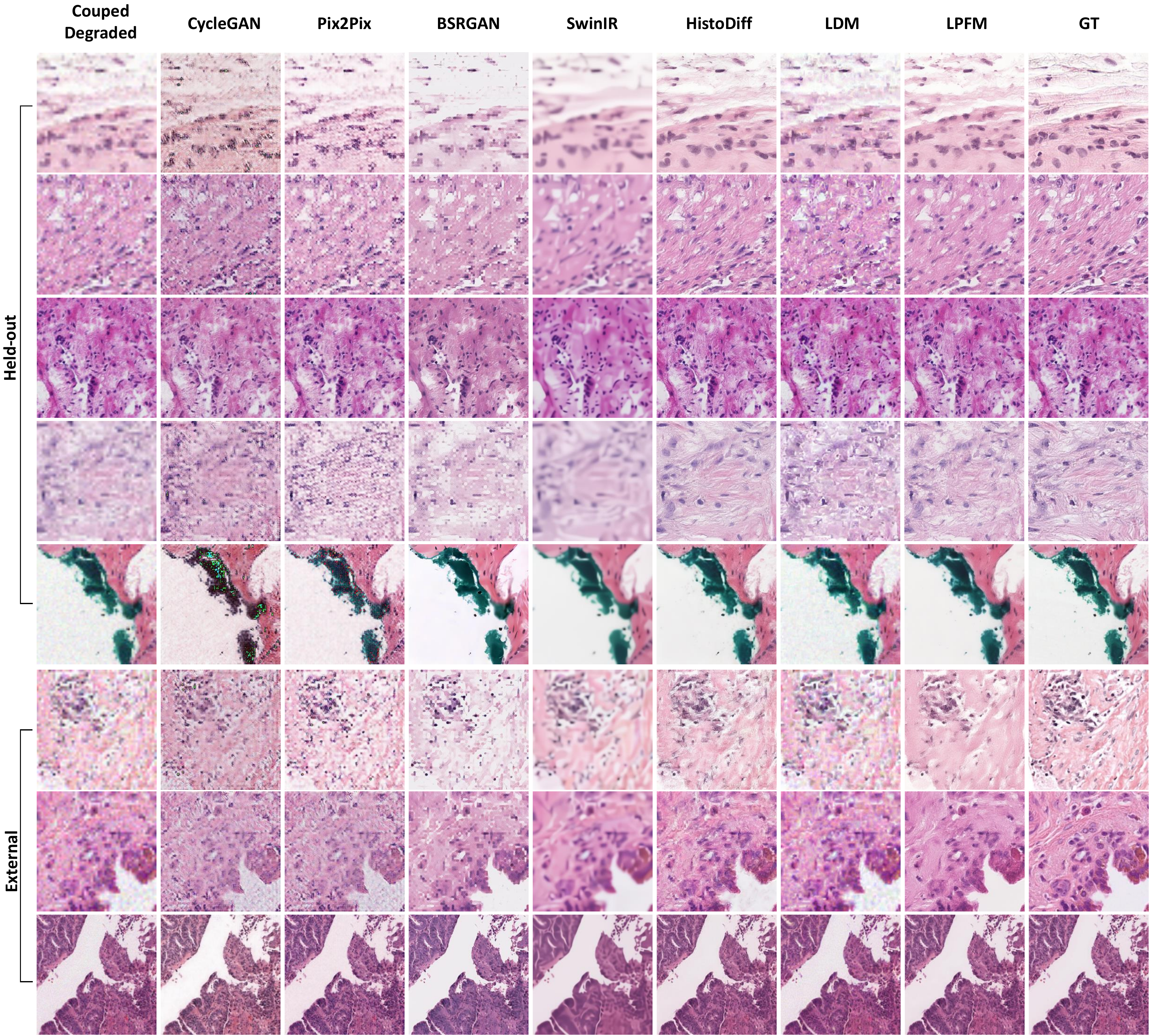}
    \caption{Extended qualitative results showcasing the restoration performance of different methods on pathology images with coupled degradations.
    }
    \label{fig: samplesCP}
\end{figure*}

We evaluated our method using a rigorous two-tier validation strategy. First, internal testing was conducted on held-out datasets from the training distribution, including CAMELYON16, PAIP2020 and PANDA. Second, external evaluation was performed on completely independent datasets, including MIDOG2022, TIGER2021 and OCELOT, to assess generalization capability. The detailed experimental results are presented in Extended Data (Tab. \ref{tab:internal_coupled_degraded_results}-\ref{tab:external_coupled_degraded_results}). More generated samples are shown in Extended Data (Fig. \ref{fig: samplesCP}).

As shown in Fig. \ref{fig: coupled_degraded_results}a-c, the quantitative results demonstrated LPFM's consistent advantages over competing approaches. In terms of PSNR, LPFM achieved a mean score of 26.15 dB, outperforming the second-best method SwinIR (24.05 dB) by a significant margin of 2.10 dB. This improvement was particularly notable in the OCELOT dataset where LPFM reached 28.20 dB, suggesting exceptional generalization capability to diverse tissue types. The SSIM results further confirmed this trend, with LPFM (0.720) substantially exceeding Pix2Pix (0.642) and other methods, indicating better structural preservation. LPFM maintained this performance advantage across all six datasets, demonstrating remarkable robustness to different degradation patterns and tissue characteristics.

LPIPS metrics, which assess perceptual quality, revealed additional insights. While HistoDiff achieved competitive LPIPS scores (0.138), LPFM demonstrated more balanced performance across all quality metrics. This suggested that while some methods might optimize for specific aspects of image quality, LPFM successfully maintained excellence in both pixel-level accuracy (PSNR) and perceptual similarity (LPIPS). The detailed experimental results are presented in Extended Data (Tab. \ref{tab:internal_coupled_degraded_results}-\ref{tab:external_coupled_degraded_results}).

Visual analysis of the results (Fig. \ref{fig: coupled_degraded_results}d-e) corroborated these quantitative findings. Compared to other methods, LPFM better preserved nuclear details and tissue architecture while more effectively suppressing artifacts. This was especially evident in the error maps where LPFM showed minimal deviation from ground truth. The intensity profile (Fig. \ref{fig: coupled_degraded_results}f-g) comparisons further demonstrated LPFM's accuracy in maintaining original image characteristics, with Pearson correlation coefficients consistently exceeding 0.9 across all test cases.

\begin{figure*}
    \centering
    \includegraphics[width=\linewidth]{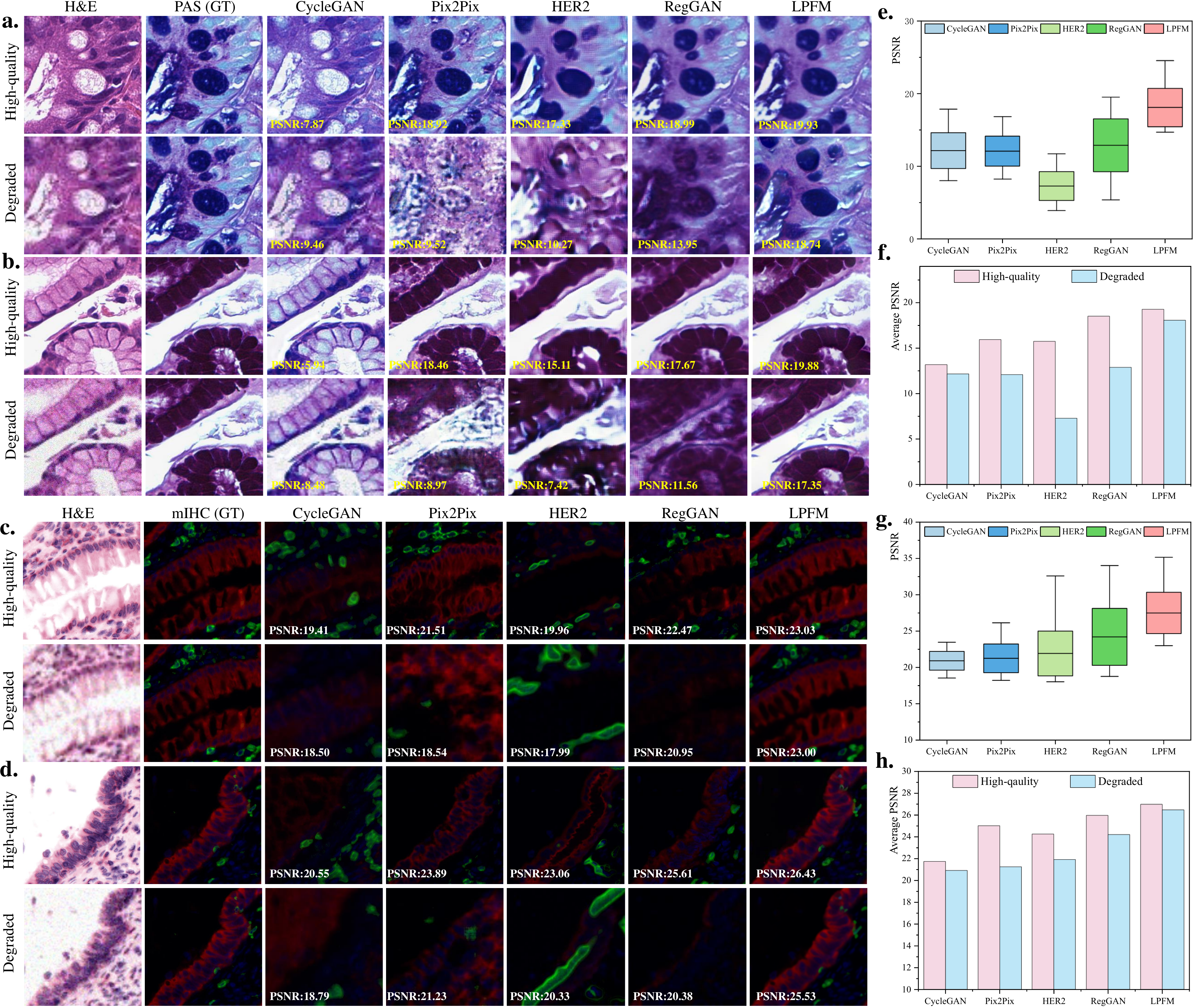}
    \caption{\textbf{Virtual staining results of pathology images with coupled degradations.}
    \textbf{a-b.} The high-quality H\&E images, degraded H\&E images, PAS ground truth (GT) images and virtually stained PAS pathology images generated by various methods on HE2PAS dataset.
    \textbf{c-d.} The high-quality H\&E images, degraded H\&E images, mIHC ground truth (GT) images and virtually stained mIHC pathology images generated by various methods on HEMIT dataset.
    \textbf{e.} PSNR performance of LFPM and various methods on HE2PAS dataset. 
    \textbf{f.} Average PSNR performance of LFPM and various methods for high-quality and degraded H\&E images on HE2PAS dataset. 
    \textbf{g.} PSNR performance of LFPM and various methods on HEMIT dataset. 
    \textbf{h.} Average PSNR performance of LFPM and various methods for high-quality and degraded H\&E images on HEMIT dataset. 
    The box limits represent the standard deviation, and the error bars indicate the 2.5\% and 97.5\% percentiles.
    }
    \label{fig: coupled_degraded_vs_results}
\end{figure*}

\subsection{Virtual Staining for Degraded Images}

%% 这里要加一个实验，首先对原图像做退化处理，施加在原图上，这里的退化可以是复合的退化问题，然后直接将之前训练的模型对原图像进行处理，得到我们虚拟染色的结果。但是我们代码层面不是输入原染色的路径的，这里需要有一个变化。初步估计是在2个数据集上做这个实验，HE2PAS和HEMIT，这两个数据集都是从HE染色开始的，所以做退化是比较容易去看的。一个从HE到PAS，一个从HE到mIHC。

%% 跟上一个小节一样，花一个Bar图就可以了。给几个sample可视化一下，给别人看看可视化的结果。

Clinical histopathology workflows frequently encounter degraded tissue specimens due to suboptimal staining, sectioning artifacts, or imaging imperfections. As illustrated in Fig. \ref{fig: coupled_degraded_vs_results}a-d, these degradations significantly impact virtual staining quality, motivating our rigorous evaluation of method robustness. Our experiments simulated realistic degradation scenarios by applying compound artifacts to H\&E images from both HE2PAS and HEMIT datasets, then processing them through pretrained models without architectural modifications.

Our experimental design simulated this clinical challenge through a two-stage process (Fig. \ref{fig: coupled_degraded_results}a-d). 
First, we applied coupled degradations (blur: $\sigma_1,  \sigma_2$=2.5, noise: $\sigma$=31, and 4$\times$ downsampling) to high-quality H\&E images from HE2PAS and HEMIT datasets.
These degraded inputs then underwent virtual staining to PAS-AB and mIHC respectively, without any intermediate restoration steps. 
This direct transformation tested the model's ability to simultaneously address staining conversion and artifact suppression.

The quantitative results in Figure~\ref{fig: coupled_degraded_vs_results}e-h demonstrated LPFM's exceptional stability across degradation conditions. On HE2PAS (Figure~\ref{fig: coupled_degraded_vs_results}e), LPFM maintained a narrow PSNR distribution (17.94-18.24 dB) for degraded inputs, outperforming RegGAN's wider range (12.68-13.09 dB) while preserving superior mean PSNR (18.09 dB vs 12.89 dB). This robustness became more pronounced in Fig. \ref{fig: coupled_degraded_vs_results}f where LPFM showed merely 6.2% performance drop from high-quality to degraded inputs, compared to RegGAN's 30.4% reduction.

The HEMIT dataset results (Figure~\ref{fig: coupled_degraded_vs_results}g) revealed an even more striking advantage - LPFM actually achieved higher PSNR on degraded inputs (26.99 dB) than on high-quality ones (26.49 dB), suggesting its unique capability to compensate for certain artifacts. As shown in Figure~\ref{fig: coupled_degraded_vs_results}h, this inverse relationship contrasted sharply with other methods' expected performance degradation, particularly HER2's 3.1\% drop. Visual comparisons in Figure~\ref{fig: coupled_degraded_vs_results}c-d confirmed that LPFM successfully suppressed noise while preserving critical histological features in mIHC staining, whereas CycleGAN introduced false positive signals and Pix2Pix lost structural detail.

% \subsection{Textual Prompt Guidance}
% % Prompt Guidance部分如何加？整体的目标是生成更清晰的结果，在这个小节里面得说明我们Prompt的作用。作用是调控模型生成，比如这里使用不同Prompt的情况下，分别做了不同类型的染色和不同类型的restore，这样说明一个模型在有交互的情况下对不同的任务都可以适配。就这样就差不多了。

% % 当然，我们的目标是得到一些清晰高质量的图像，但是。具体来说，这里举例说可以分别对举例对虚拟染色进行不同染色的操作，例如对同一张HE图像，生成对应的HE2PAS图像，或者生成对应的HEMIT图像。对于退化复原问题，

% % 老师说这里放在ablation study。然后需要说明，如果输入的是高质量图像，也可以原封不动得到对应的高质量图像，解释原因并且说明为什么可以做到。

% \subsection{High-level Downstream Evaluation}
% High-level downstream task

% 这个不是必须的，老师说这个不一定需要，因为没有很高的提升， 不一定有意义。所以这个可以放在Discussion来说明，说明未来的一些工作或者可以帮助到的一些方面。

\section{Discussion}
% 这里需要改一下，突出重点，给出几个小标题吧，
% textual prompt guidance 
% effectiveness of contrastive learning 
% effectiveness of image refinement 
% Clinical Implications
% Future Work.

The development of the Low-level Pathology Foundation Model (LPFM) represents a significant advancement in computational pathology by unifying diverse image restoration and virtual staining tasks within a single, adaptable framework. Our comprehensive evaluation across 66 distinct tasks demonstrates that this unified approach not only matches but frequently surpasses the performance of specialized models while offering unprecedented flexibility for clinical and research applications. 
The success of LPFM stems from its integration of several key innovations. The contrastive pre-training strategy enables the model to learn robust, stain-invariant feature representations that generalize across diverse tissue types and degradation patterns. This is complemented by a unified conditional diffusion framework that dynamically adapts to specific tasks through textual prompts, allowing clinicians to guide the enhancement process based on diagnostic priorities. Training on an extensive dataset encompassing 34 tissue types and 5 staining protocols further ensures broad applicability, from routine H\&E analysis to specialized staining techniques. Notably, LPFM excels in handling coupled degradations, where it achieves a 2.10 dB PSNR improvement over leading specialized methods. This capability is particularly valuable given that clinical images often exhibit multiple concurrent artifacts that require simultaneous correction.

\begin{figure*}[t]
    \centering
    \includegraphics[width=0.98\linewidth]{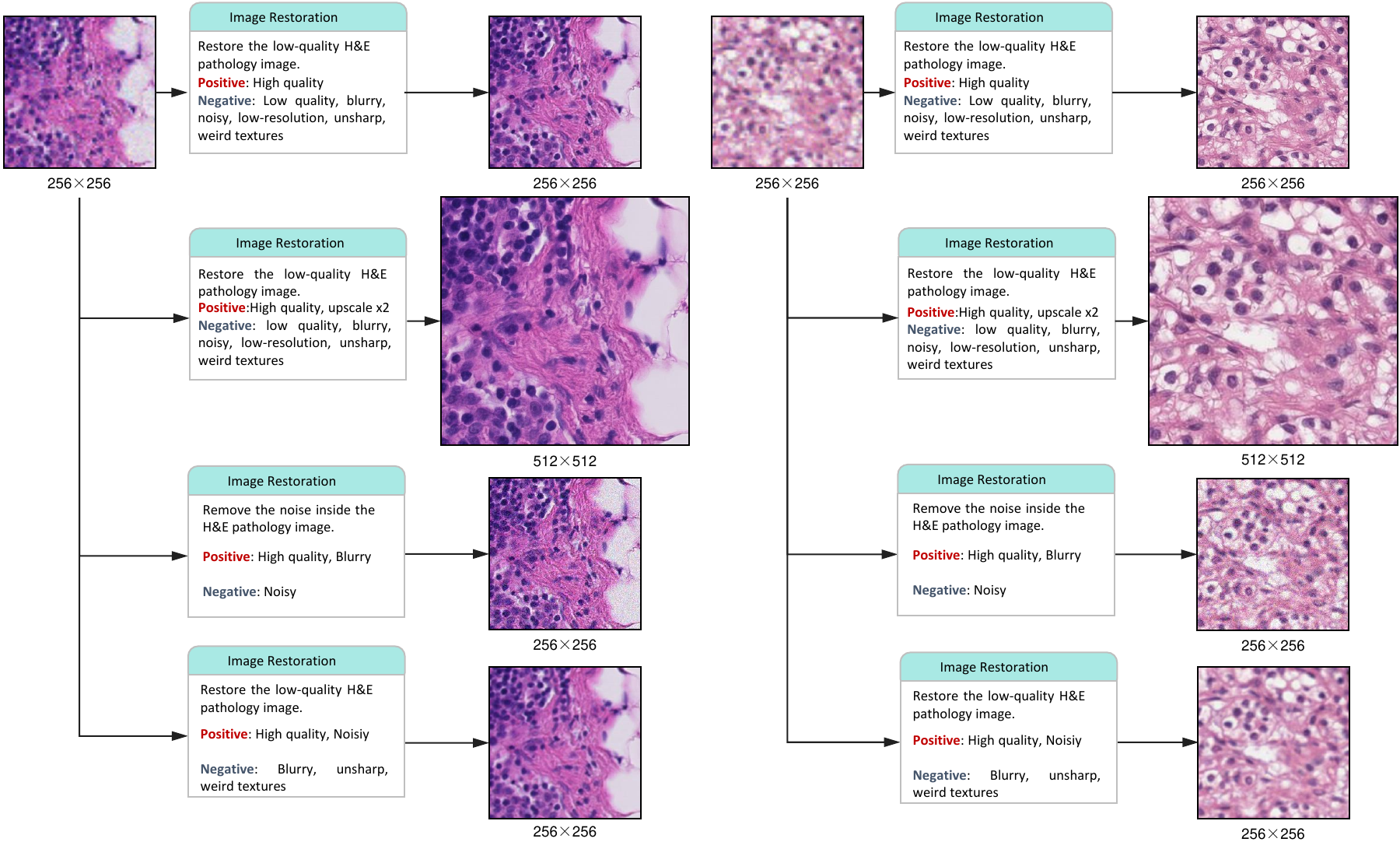}
    \caption{H\&E pathology image restoration examples with varying textual prompt guidance. 
    }
    \label{fig: PromptGuidance_Restoration}
\end{figure*}

\begin{figure*}[t]
    \centering
    \includegraphics[width=0.98\linewidth]{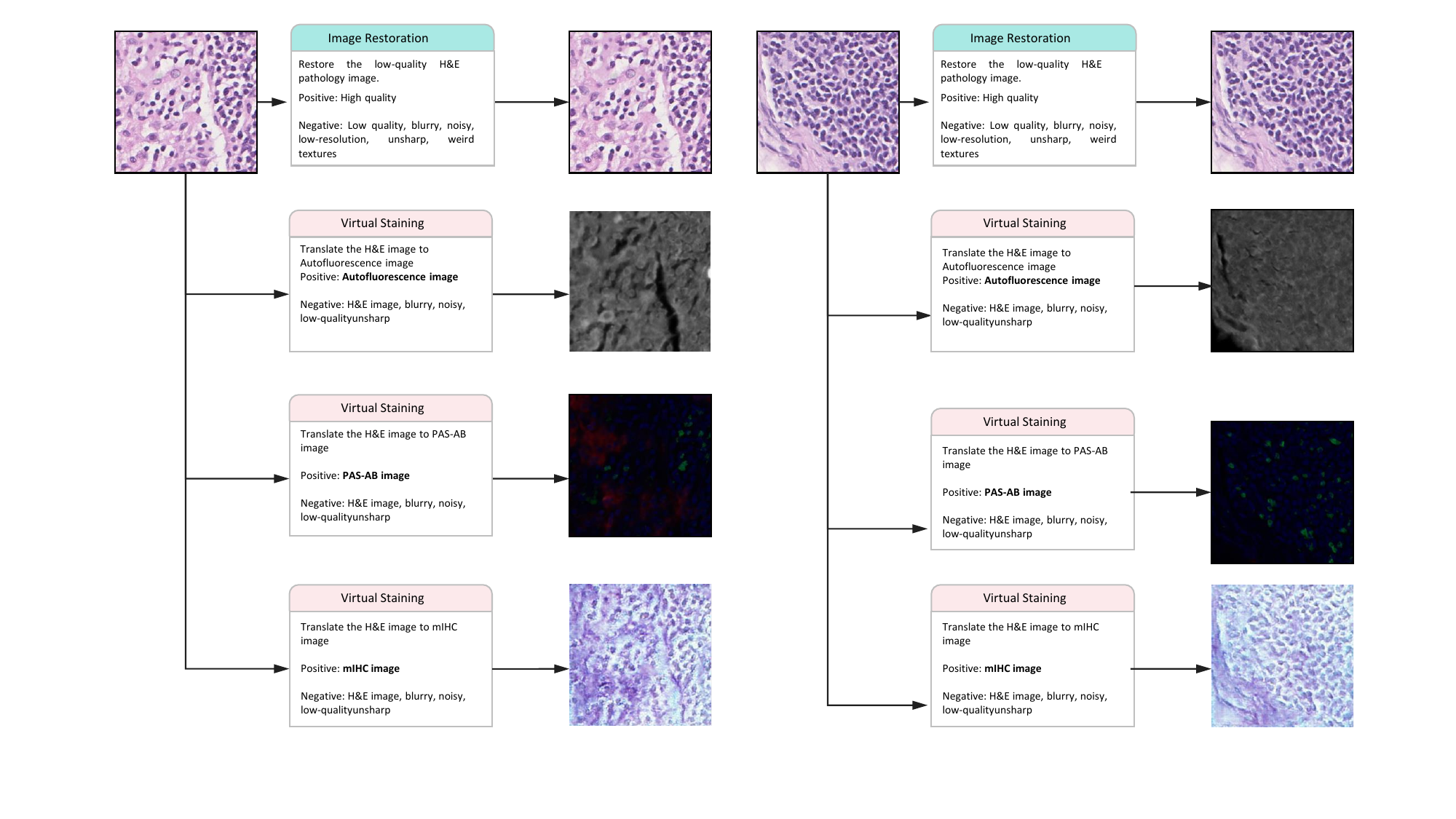}
    \caption{Pathology image virtually staining examples with varying textual prompt guidance. 
    }
    \label{fig: PromptGuidance_Virtualstaining}
\end{figure*}

\noindent \textbf{Textual Prompt Guidance.}
The prompt-guided architecture introduces a novel level of interactivity to pathology image processing. By allowing users to specify enhancement priorities through natural language instructions, the system adapts to diverse diagnostic needs without requiring architectural modifications. 
As shown in Fig. \ref{fig: PromptGuidance_Restoration} and \ref{fig: PromptGuidance_Virtualstaining}, the system dynamically adapts its image restoration or virtual staining targets based on natural language instructions without requiring architectural modifications.This flexibility addresses a major limitation of conventional approaches that enforce rigid processing pipelines, potentially enabling more personalized and context-aware low-level tasks.

\begin{figure*}[t]
    \centering
    \includegraphics[width=\linewidth]{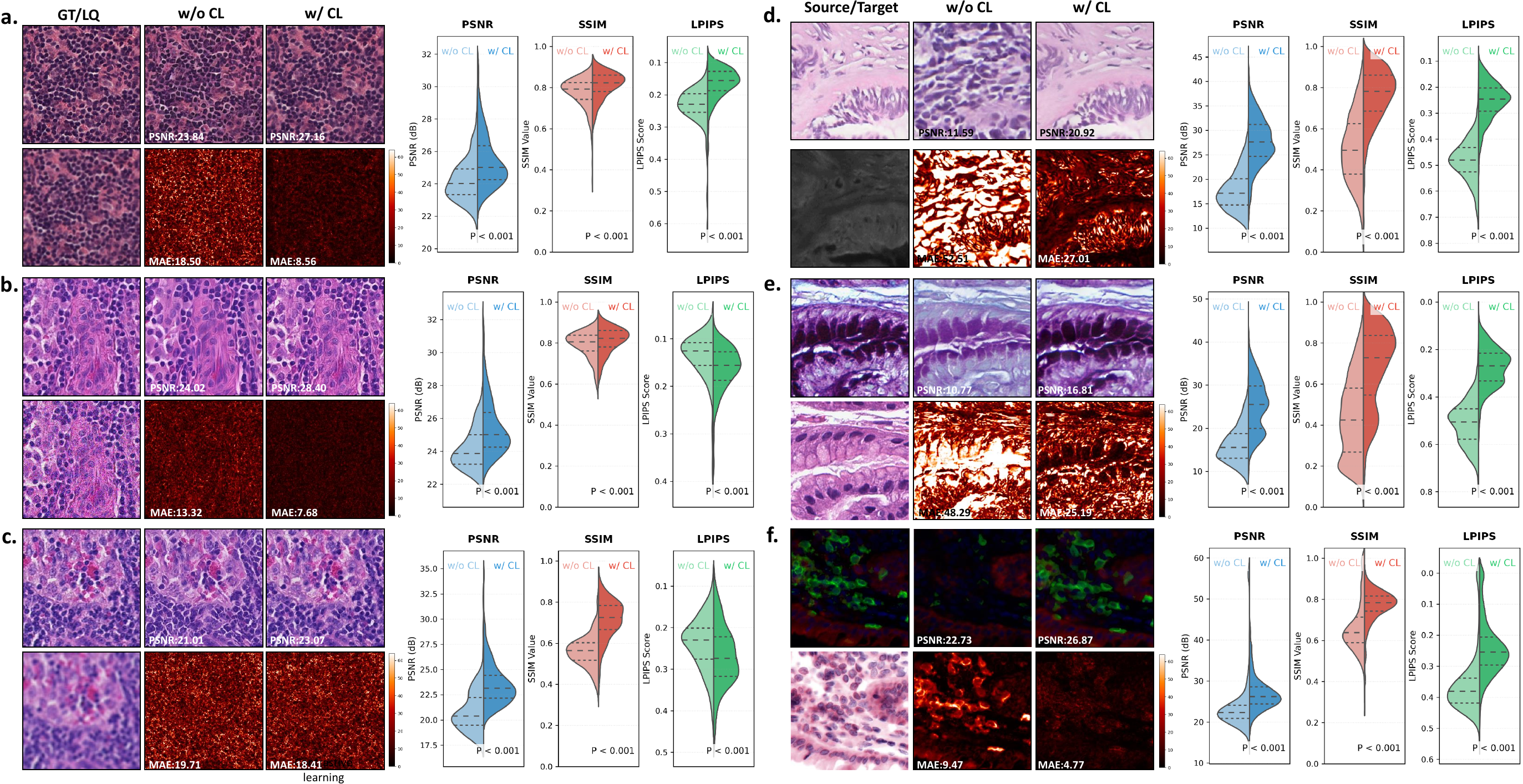}
    \caption{Effectiveness of autoencoder pretraining via contrastive learning (CL) for pathology image restoration and virtually staining. 
    \textbf{a-c.} The high-quality ground truth pathology images, degraded low-quality pathology images (a. low resolution, b. noise, c. blur), mean average error (MAE) images and restored pathology images generated by LPFM with or withour CL. Results of PSNR, SSIM and LPIPS of LPFM with or without CL on CAMELYON16 are shown in the right figures.
    \textbf{d-f.} The source/target stained pathology images, mean average error (MAE) images and virtually stained pathology images generated by LPFM with or withour CL. Results of PSNR, SSIM and LPIPS of LPFM with or without CL on different stained dataset (d. AF2HE, e. HE2PAS, f. HEMIT) are shown in the right figures. The horizontal dashed lines indicate the mean metrics. The dotted lines indicate 25\% and 75\% percentile. 
    }
    \label{fig: woCL}
\end{figure*}

\noindent \textbf{Effectiveness of Contrastive Pre-training.}
As demonstrated in Fig. \ref{fig: woCL}, contrastive learning provides fundamental improvements to the model's ability to handle both image restoration and virtual staining tasks. The visual comparisons clearly show that the contrastive pre-trained version better preserves fine cellular structures and tissue morphology compared to the non-contrastive baseline. This improvement is particularly evident in the mean average error (MAE) mapswhere the contrastive approach shows significantly reduced errors in diagnostically important regions. The quantitative results across all three evaluation metrics consistently confirm these visual observations, demonstrating that contrastive learning enables more robust feature representations for diverse pathology image processing tasks.

\begin{figure*}[t]
    \centering
    \includegraphics[width=\linewidth]{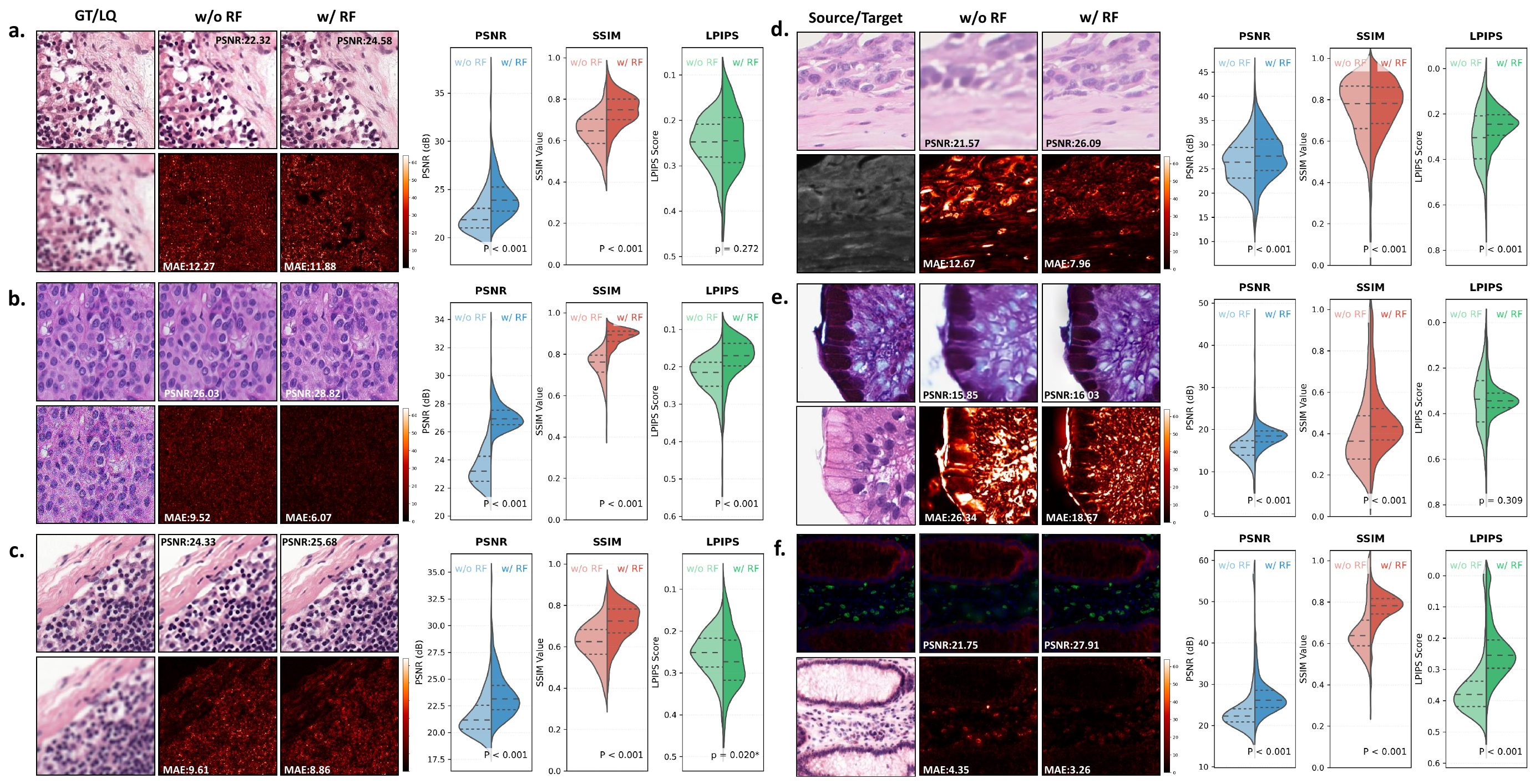}
    \caption{Effectiveness of image refinement (RF) via conditional diffusion model for pathology image restoration and virtually staining. 
    \textbf{a-c.} The high-quality ground truth pathology images, degraded low-quality pathology images (a. low resolution, b. noise, c. blur), mean average error (MAE) images and restored pathology images generated by LPFM with or withour RF. Results of PSNR, SSIM and LPIPS of LPFM with or without RF on CAMELYON16 are shown in the right figures.
    \textbf{d-f.} The source/target stained pathology images, mean average error (MAE) images and virtually stained pathology images generated by LPFM with or withour RF. Results of PSNR, SSIM and LPIPS of LPFM with or without RF on different stained dataset (d. AF2HE, e. HE2PAS, f. HEMIT) are shown in the right figures. The horizontal dashed lines indicate the mean metri  0=cs. The dotted lines indicate 25\% and 75\% percentile. 
    }
    \label{fig: woRF}
\end{figure*}

\noindent \textbf{Effectiveness of Pathology Image Refinement.}
The refinement module plays a crucial role in enhancing output quality, as shown in Fig. \ref{fig: woRF}. The conditional diffusion-based refinement effectively suppresses artifacts while preserving critical diagnostic features in both restoration and virtual staining tasks. Visual examination of the results reveals that the refined outputs maintain better structural consistency with the ground truth, particularly in challenging regions such as tissue boundaries and cellular details. The quantitative metrics consistently support these observations, with the refined versions showing superior performance across all evaluation criteria. The improvement is maintained across different types of image degradation and staining transformations, demonstrating the general applicability of the refinement approach.

\noindent \textbf{Potential Limitation.} 
Despite these advances, certain limitations must be acknowledged. While LPFM generalizes well across multiple scanners and staining protocols, its performance may vary when confronted with radically novel imaging modalities not represented in the training data. Additionally, like all generative models, there remains a small risk of introducing plausible but artifactual features in severely degraded inputs. These observations highlight important directions for future research, including extension to 3D pathology data, development of interactive refinement mechanisms, and creation of explainability tools to clarify enhancement decisions.

\noindent \textbf{Clinical Implications.} From a clinical perspective, LPFM's virtual staining capabilities could transform diagnostic workflows. The model's ability to generate high-quality PAS-AB and mIHC stains from H\&E images with consistent structural preservation offers practical advantages, including reduced reagent costs, faster turnaround times, and conservation of precious biopsy material. These benefits would be especially impactful for rare specimens or when additional stains are needed retrospectively. While our quantitative and qualitative analyses confirm that virtually stained images maintain diagnostically critical features such as nuclear membranes and glycoprotein distributions, further clinical validation studies will be essential to assess diagnostic concordance with conventional staining methods.
The broader implications of this work extend beyond immediate clinical applications. By demonstrating that a single model can excel at multiple low-level vision tasks while maintaining diagnostic reliability, LPFM reduces the computational overhead associated with deploying multiple specialized solutions in pathology workflows. This unification could improve the interoperability of computational tools while making advanced image enhancement more accessible to resource-limited settings. Moreover, the prompt-based control mechanism establishes a template for collaborative human-AI interaction in medical image interpretation where clinician expertise guides algorithmic processing to align with diagnostic requirements. Some examples of prompts controlling the generation are shown in Extended Data (Fig. \ref{fig: PromptGuidance_Restoration}-\ref{fig: PromptGuidance_Virtualstaining}).

\noindent \textbf{Future works.} 
Two critical directions warrant further investigation. First, extending LPFM to 3D pathology data represents a natural evolution, as volumetric imaging becomes increasingly important for comprehensive tissue analysis. This extension would enable artifact correction and virtual staining across z-stack acquisitions while presenting new computational challenges in processing high-dimensional pathology data. Second, large-scale clinical validation studies are essential to rigorously quantify diagnostic concordance between enhanced and ground-truth images across diverse disease subtypes and staining protocols, particularly focusing on whether model-generated features maintain diagnostic fidelity in critical regions like tumor margins or micrometastases.

In conclusion, LPFM advances the field of computational pathology by providing a versatile, high-performance solution for image quality enhancement that addresses both common and challenging real-world scenarios. The model's ability to handle diverse tasks through a unified framework—while maintaining or improving upon the performance of specialized alternatives—suggests that foundation models can achieve transformative impact in medical imaging, much as they have in other domains. As the field progresses, the integration of such models into clinical workflows, coupled with ongoing technical refinements and rigorous validation, promises to enhance the accuracy, efficiency, and accessibility of pathological diagnosis.

\section{Methods}\label{methods}

% 仍需要做的工作
% 1. Pretraining datasets 需要加，说明我们用于预训练的数据
% 2. 跟本子里面的内容保持一致，除了prompt控制调控方向和生成的质量
% Methods部分和之前写本子的时候保持一致

\subsection{Preparation Process}
To ensure the robustness and generalizability of our model on image restoration and virtual staining tasks, we collected a comprehensive, multi-source dataset encompassing diverse tissue types and staining protocols. Our dataset includes H\&E (Hematoxylin and Eosin), IHC (Immunohistochemistry), mIHC (multiplex immunohistochemistry) images, and PAS (Periodic Acid-Schiff) stained slides from multiple organs (e.g., liver, kidney, breast, and lung) (Extended Data Tab. \ref{tab:data_primary_site}).

\subsubsection{Whole Slide Image Tiling and Stitching}\label{wsi_preprocess}
Due to the extremely high resolution of WSIs (often exceeding 100,000 $\times$ 100,000 pixels), direct processing is computationally infeasible. Thus, we partition each WSI into smaller, manageable patches of size 256 $\times$ 256 pixels with 32 pixel overlapping, ensuring sufficient spatial context for image restoration and virtual staining while maintaining computational efficiency.
After image restoration and virtual staining, the processed patches are reassembled into a complete WSI using grid-based stitching with 32-pixel overlapping boundaries to eliminate seam artifacts. Bilinear interpolation is applied to ensure smooth transitions between adjacent patches.

\subsubsection{Degradation Simulation for Restoration Tasks}\label{degradation_approach}
To generate paired training data for our image restoration tasks, we simulate three clinically relevant degradation types through carefully designed transformations of high-quality WSIs. Each degradation method was developed in consultation with pathologists to ensure biological plausibility. We mainly analyze three typical degradation types in pathology images, including low-resolution, blurry and noisy problems, which commonly occur during whole-slide image acquisition due to various factors such as optical limitations, tissue preparation artifacts, and scanning imperfections. Below we detail the specific implementations for each degradation type. 

\noindent \textbf{Low Resolution}: 
In whole-slide imaging systems, resampling operations play a critical role in generating multi-resolution pyramids for pathological analysis. Our model incorporates three clinically validated interpolation methods (area-based, bilinear, and bicubic) that reflect the resampling algorithms used in commercial whole-slide scanners. Area-based interpolation best preserves nuclear morphology and intensity distributions, while bilinear maintains smooth tissue transitions and bicubic captures fine chromatin textures, though it may introduce slight edge enhancements. We intentionally exclude nearest-neighbor interpolation due to its tendency to create jagged nuclear borders and artificial discontinuities in tissue architecture that could mimic pathological features. During training, we randomly select among the three approved methods to ensure robustness across different scanner implementations. This approach was validated through consultation with pathologists, who confirmed it successfully maintains three key diagnostic features: nuclear membrane integrity for tumor grading, stromal texture for invasion assessment, and chromatin patterns for molecular characterization. Importantly, the method realistically simulates the multi-resolution acquisition process inherent to digital pathology workflows.

\noindent \textbf{Image Blur}: 
In computational pathology, blur artifacts commonly arise from optical defocus, tissue sectioning imperfections, and scanner vibrations. We model these effects using anisotropic Gaussian kernels that account for the directional variability observed in pathological imaging. The blur kernel is defined as:

\begin{equation}
\boldsymbol{k}(i,j) = \frac{1}{N} \exp\left(-\frac{1}{2}\boldsymbol{C}^T\boldsymbol{\Sigma}^{-1}\boldsymbol{C}\right), \quad \boldsymbol{C}=[i,j]^T
\end{equation}
where the covariance matrix $\Sigma$ controls the blur's directionality; $C$ is the spatial coordinates; $N$ is the normalization constant. The covariance matrix could be further represented as follows:
\begin{equation}
    \begin{aligned}
\boldsymbol{\Sigma} & =\boldsymbol{R}\left[\begin{array}{cc}
\sigma_{1}^{2} & 0 \\
0 & \sigma_{2}^{2}
\end{array}\right] \boldsymbol{R}^{T}, \quad(\boldsymbol{R} \text { is the rotation matrix }) \\
& =\left[\begin{array}{cc}
\cos \theta & -\sin \theta \\
\sin \theta & \cos \theta
\end{array}\right]\left[\begin{array}{cc}
\sigma_{1}^{2} & 0 \\
0 & \sigma_{2}^{2}
\end{array}\right]\left[\begin{array}{cc}
\cos \theta & \sin \theta \\
-\sin \theta & \cos \theta
\end{array}\right],
\end{aligned}
\end{equation}
This formulation captures both isotropic defocus blur (when $\sigma_{1} = \sigma_{2}$) and anisotropic artifacts from scanner motion or uneven tissue surfaces (when $\sigma_{1} \neq \sigma_{2}$). The rotation matrix accounts for directional effects commonly seen in whole-slide scanning. We exclude unrealistically sharp kernels that could distort nuclear morphology and other diagnostically important features.

\noindent \textbf{Image Noise}: 
Noise artifacts primarily originate from two distinct physical processes: electronic sensor noise and quantum photon fluctuations. We model these phenomena using a composite noise formulation that accounts for the unique characteristics of pathological imaging. Additive Gaussian noise captures the electronic readout noise inherent in CCD/CMOS sensors, with its intensity modulated by the standard deviation $\sigma$ of the normal distribution $N(0, \sigma^2)$. This noise component manifests as color noise (when independently sampled across RGB channels), reflecting different scanner architectures.

Poisson noise models the fundamental quantum limitations of photon detectionwhere the variance scales linearly with signal intensity according to the Poisson distribution $P(\lambda I)$. This noise source is particularly relevant in high-magnification imaging of weakly stained regionswhere photon counts are inherently limited. The combination of these noise types effectively reproduces the characteristic graininess observed in low-light conditions while maintaining the structural integrity of diagnostically critical features such as nuclear membranes and chromatin patterns.
Therefore, the noisy image can be represents as:
\begin{equation}
\begin{aligned}
I_{\text{noisy}}(x,y) = I_{\text{clean}}(x,y) + \mathcal{N}(0,\sigma^2) + \mathcal{P}(\lambda I_{\text{clean}}(x,y)) - \lambda I_{\text{clean}}(x,y)
\end{aligned}
\end{equation}
where $I_{clean}(x,y)$ represents the noise-free image intensity at pixel $(x,y)$, $\mathcal{N}(0,\sigma^2)$ denotes additive Gaussian noise.  $\mathcal{P}(\lambda I_{clean}(x,y))$ models Poisson-distributed quantum noise:
\begin{equation}
P(k) = \frac{(\lambda I)^k e^{-\lambda I}}{k!}, \quad k=0,1,2,...
\end{equation}

Our implementation preserves the Poisson noise's signal-dependent nature while avoiding artificial amplification of staining variations, ensuring biologically plausible noise characteristics throughout the dynamic range of pathological specimens.

% \subsubsection{Image Registration for Virtual Staining Tasks}

\subsection{Network Architecture}
% 这里应该包括2个部分，一个是预训练，一个是微调
% 预训练包括3个，一个时feature encoder 的预训练，一个是diffusion的预训练，还有就是最后的推理阶段，在预训练阶段中，需要加入额外的Promt作为指导，来控制生成配对图像的方向，比如在预训练过程中，有一些人物的目标是复原，有一些人物的目标是做虚拟染色，或者是复合退化和特定染色，而在生成过程中，我们的目标基本上都是得到目标染色的更清晰的图像，在说明这个过程中需要结合前面的results部分一起说明，来体现我们这个设计中Prompt对于预训练的帮助。其中需要给出
% 下游阶段微调阶段，说明在这个阶段中，我们是如何将预训练的结构用来微调的，为什么微调阶段和预训练阶段要分开来处理，在微调阶段需要那些额外的不同于预训练的部分
% 结合Diffusion 的推理过程是什么样的，推理阶段和预训练阶段相比有什么不同，推理过程的参数设置是什么样的，

The objective of this work is to develop a unified low-level pathology foundation model for image restoration and virtual staining. To ensure robust interactivity and enable precise control across multiple tasks, our model employs a prompt-based conditioning mechanism, which dynamically guides the generation process toward the desired output modality (e.g., restoration, virtual staining, or coupled degradation reversal). 

Our framework employs a two-stage training approach to achieve high-fidelity pathology image generation through progressive refinement. In the first stage, a contrastive learning-based autoencoder learns to extract consistent low-level features and produce coarse restorations guided by task-specific prompts. Building upon this foundation, the second stage trains a denoising diffusion model that takes both the coarse reconstruction and prompt embedding as inputs to generate detailed, high-quality outputs. The diffusion model progressively refines the image through an iterative denoising process, while maintaining anatomical consistency from the coarse input and adhering to task-specific requirements through prompt conditioning. This hierarchical approach allows our unified model to first capture global structural information and then synthesize precise local details, enabling flexible performance across diverse tasks, including image restoration and virtual staining, without requiring architectural modifications. The diffusion model's conditioning on both the initial reconstruction and textual prompts ensures that the final output not only preserves the input's structural integrity but also accurately reflects the desired transformation specified by the prompt, whether it involves stain conversion, artifact removal, or resolution enhancement.

During the inference stage, our unified low-level pathology foundation model takes a user-defined prompt and noisy image as input to generate high-quality image through a controlled reverse diffusion process. The pretrained encoder first processes the textual or embedding-based prompt to extract task-specific conditioning signals, while the diffusion model progressively denoises the initial random noise across multiple timesteps. This iterative refinement enables the synthesis of both structurally accurate and task-aligned results - whether restoring low-quality H\&E pathology images (prompt: "Obtain high-quality H\&E pathology image") or generating specific stains (prompt: "Synthesize H\&E staining") - while maintaining tissue-level consistency. 

\noindent \textbf{Contrastive Pretrainig for Coarse Restoration }

In the pre-training phase, we aim to pretrain a low-level autoencoder that can extract consistent low-level features and produce coarse restorations guided by task-specific prompts.  Following LDM\cite{rombach2022high} and CLIP\cite{radford2021learning}, we pretrain the KL-Autoencoder as the low-level pathology image autoencoder to generate the coarse restorations and directly use the CLIP\cite{radford2021learning} text encoder to obtain the textual features for guidance. 

\begin{figure*}[t]
    \centering
    \includegraphics[width=0.96\linewidth]{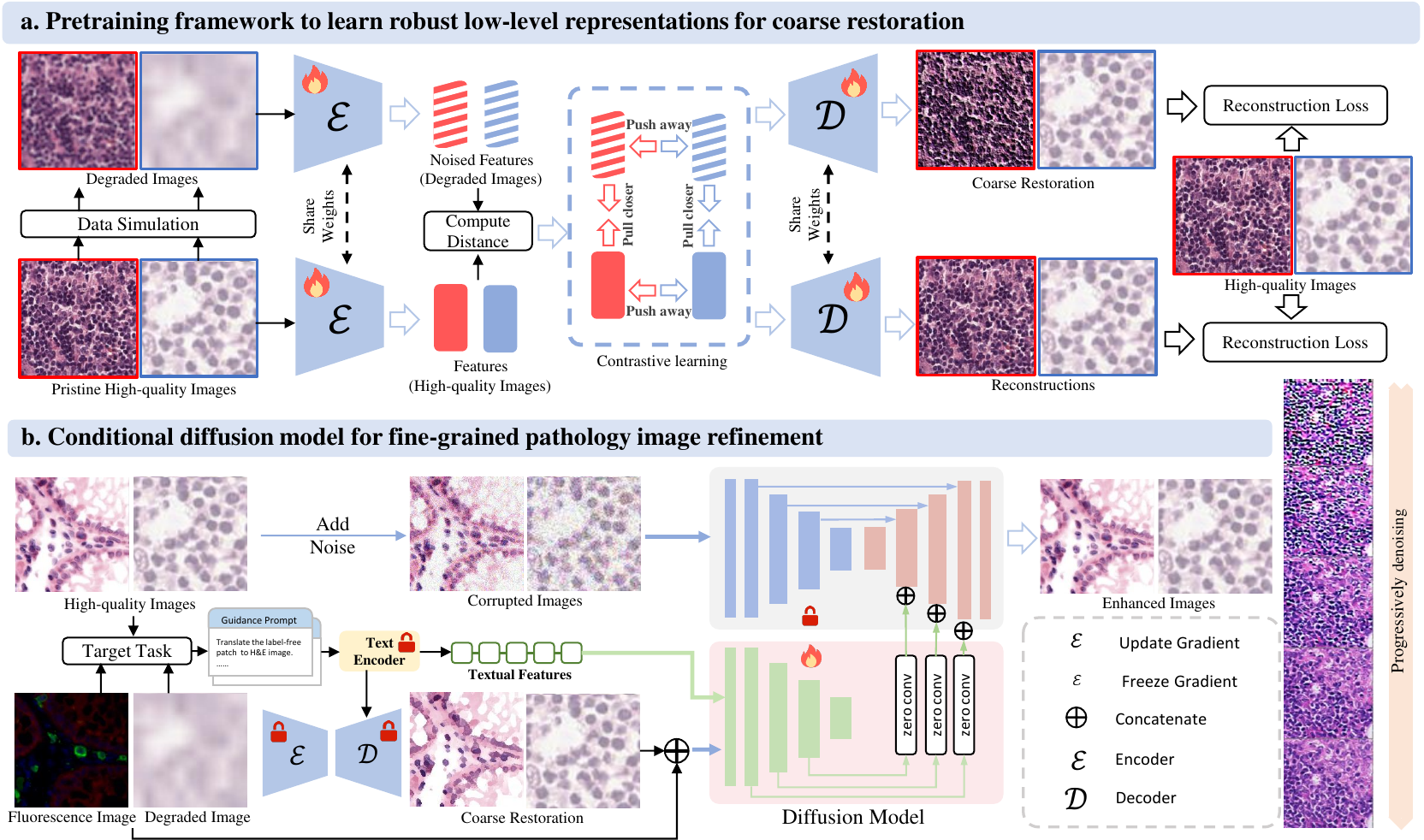}
    \caption{The pipeline of unified low-level pathology foundation model in the training stage. \textbf{a.} We propose a pretraining framework that learns degradation-robust representation through contrastive learning and pixel-wise reconstruction, enabling coarse restoration of images with coupled degradations. \textbf{b.} We propose a conditional diffusion model that improves image quality through a guided denoising process, utilizing both the coarse restorations and textual prompts as conditional inputs 
    }
    \label{fig: pipeline_training}
\end{figure*}

Given the WSIs, we firstly pre-process the WSIs into 256$\times$256 pathology patches as early mentioned in sec. \ref{wsi_preprocess}.  
As shown in Extended Data Fig. \ref{fig: pipeline_training}a, we adopt a unified training paradigm for low-level pathology tasks by leveraging contrastive learning to capture shared feature representations across different staining protocols and image quality levels. This approach enables the model to learn robust latent embeddings that are invariant to degradation types and staining variations in histopathology images. The contrastive loss operates in the latent space, pulling together features from different views (e.g., degraded/restored or differently stained versions) of the same tissue while pushing apart features from different tissue samples:
\begin{equation}
\mathcal{L}_{cont} = -\mathbb{E}_{x,x^+}\left[\log\frac{\exp(\mathcal{E}(x)^T\mathcal{E}(x^+)/\tau)}{\sum_{x^-}\exp(\mathcal{E}(x)^T\mathcal{E}(x^-)/\tau)}\right]
\end{equation}
where $x$ and $x^+$ are positive pairs (different views of same tissue), $x^-$ are negative samples (different tissues), $\mathcal{E}(\cdot)$ is the feature encoder, and $\tau$ is a temperature parameter.

For high-quality reference images in restoration tasks and target-stain images in virtual staining tasks, the KL-autoencoder is optimized to reconstruct the original input through the objective: 
\begin{equation}
    \mathcal{L}_{recon} = \mathbb{E}_{x\sim p(x)}[|x - \mathcal{D}(\mathcal{E}(x))|_1]
\end{equation}
where $\mathcal{E}$ and $\mathcal{D}$ represent the encoder and decoder respectively. 
Simultaneously, when processing degraded inputs for restoration or source-stain images for virtual staining, the model is trained to generate enhanced outputs that approximate the target distribution using a composite loss function:
\begin{equation}
    \mathcal{L}_{enhance} = \mathbb{E}_{x\sim p(x)}[|x - \mathcal{D}(\mathcal{E}(x_{d}))|_1]
\end{equation}
where $x_{p}$ represent the paired degraded or source-stained pathology images. 
The reconstruction loss ensures pixel-level accuracy by minimizing absolute differences between generated and target images. However, pixel-wise losses alone often produce blurry results, so we incorporate a perceptual loss (LPIPS) \cite{johnson2016perceptual} that operates in the feature space of a pretrained VGG network \cite{simonyan2014very}:
\begin{equation}
\small
    \mathcal{L}_{\text{perceptual}} = \mathbb{E}_{x,y}\left[\sum_{l}\frac{1}{H_lW_l}\|\psi_l(y) - \psi_l(\mathcal{D}(\mathcal{E}(x)))\|_1\right]
\end{equation}
where $\psi_l(\cdot)$ denotes features from layer $l$ of a pretrained VGG network, and $H_l/W_l$ are spatial dimensions of the feature maps.

To achieve more realistic quality for pathology images, we employ an adversarial loss that aligns the generated image distribution with original pathology images through a discriminator.
\begin{equation}
\mathcal{L}_{\text{adv}} = \mathbb{E}_y[\log \mathcal{D}(y)] + \mathbb{E}_x[\log(1 - D(\mathcal{D}(\mathcal{E}(x)))]
\end{equation}
where discriminator $D$ is trained to distinguish between real ($y$) and generated ($D(E(x))$) images.
Therefore, the final pretraining loss can be denoted as:
\begin{equation}
    \mathcal{L} = \mathcal{L}_{recon} + \mathcal{L}_{enhance} + \mathcal{L}_{cont} + \mathcal{L}_{\text{adv}} + \mathcal{L}_{\text{perceptual}} 
\end{equation}

The textual prompts encoded by the CLIP text encoder are injected into the encoder through cross-attention layers, providing explicit guidance to steer the reconstruction toward the desired output. This integrated approach ensures the autoencoder learns semantically meaningful representations while maintaining flexibility across diverse pathology image processing tasks.

\noindent \textbf{Conditional Diffusion for Image Refinement }:

Building upon the pretrained autoencoder's coarse restorations, we implement a conditional diffusion model to recover fine pathological details through noise-to-image synthesis. As shown in Extended Data Fig. \ref{fig: pipeline_training}b, we aim to remove the additive noises in the pathology images with the coarse image generation and textual prompts.

Our framework builds upon diffusion probabilistic models, which learn a target data distribution $p(x)$ through iterative denoising of normally distributed variables. 
These models formulate the generation process as the reverse of a fixed Markov chain spanning $T$ timesteps, effectively implementing a sequence of denoising autoencoders $\epsilon_\theta(x_t,t); t=1\dots T$. 
Following the standard controllable diffusion framework\cite{zhang2023adding}, we train a U-Net to iteratively denoise corrupted versions of high-quality pathology images. For a target image $x_0$ progressively noised to $x_t$ at timestep $t$, the model is trained to predict and remove noise from corrupted versions $x_t$ of input images $x$, following the reweighted variational lower bound objective:
\begin{equation}
\mathcal{L}_{\text{DM}} = \mathbb{E}_{x,\epsilon\sim\mathcal{N}(0,1),t}\left[\|\epsilon - \epsilon_\theta(x_t,t)\|^2_2\right]
\label{Ldm}
\end{equation}
with t uniformly sampled from ${1, . . . , T }$.

In our diffusion model development, we adopt a two-phase training strategy to ensure robust noise modeling and precise conditional control. First, we pretrain a standard denoising diffusion model without any conditional inputs, optimizing the baseline objective from Eq. \ref{Ldm}. This initial phase establishes fundamental denoising capabilities for pathology images. After convergence, we freeze these parameters and introduce a trainable controllable module that shares the U-Net\cite{ronneberger2015u} encoder with the fixed diffusion model. The complete architecture then processes both the coarse reconstruction $z$ and textual prompt embedding $c$ through and minimize the joint optimization:
\begin{equation}
\mathcal{L}_{\text{cond}} = \mathbb{E}_{x,\epsilon,c,t}\left[\|\epsilon - \epsilon_\theta(x_t,t,z,c)\|^2_2\right]
\end{equation}
where $\epsilon_\theta(x_t,t,z,c)$ represents the predicted noised inside the image from the diffusion model.
It should be noted that all the denoised process could be directly applied in the latent space of the pretrained autoencoder from the first stage. This latent diffusion approach can reduce computational costs as the dimensionality of the latent space is much lower than that of the original image space.

\noindent \textbf{Inference Stage }

\begin{figure*}[t]
    \centering
    \includegraphics[width=0.96\linewidth]{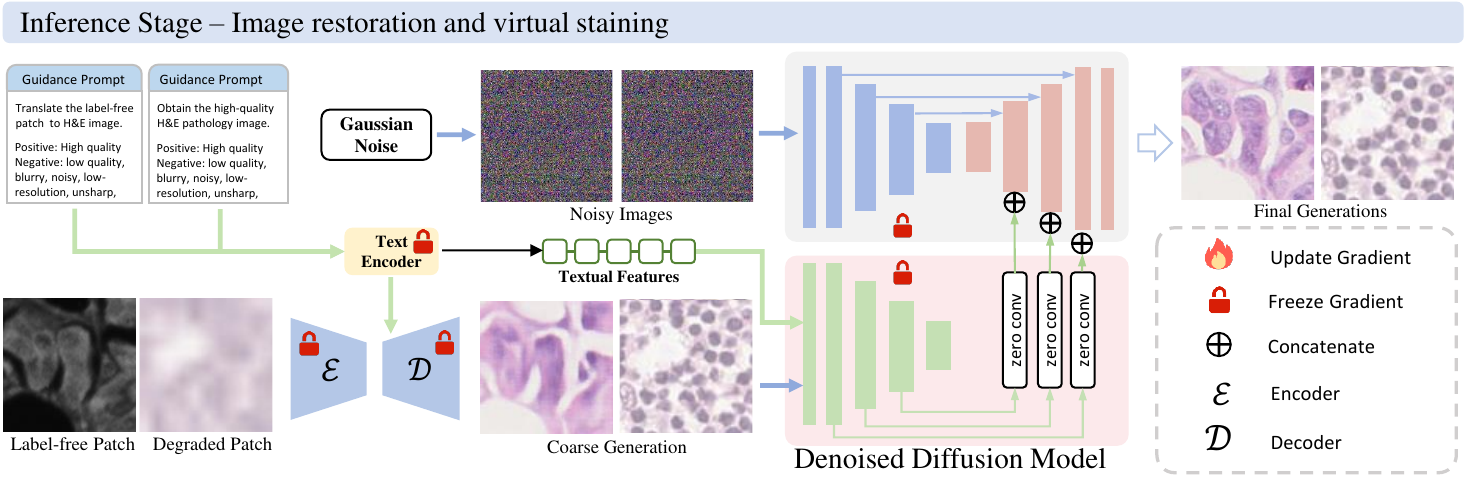}
    \caption{The inference pipeline of unified low-level pathology foundation model for image restoration and virtual staining. }
    \label{fig: pipeline_inference}
\end{figure*}

In the inference stage, our model synthesizes high-quality pathology images through an iterative denoising process that combines coarse image reconstructions and textual prompts in Fig. \ref{fig: pipeline_inference}. The generation begins with pure Gaussian noise $x_T \sim \mathcal{N}(0,\mathbf{I})$ and progressively refines it over T timesteps using the trained diffusion model conditioned on both the encoded coarse image $z$ and the prompt embedding $c$. At each timestep t, the model predicts the noise component using:
\begin{equation}
\hat{\epsilon}_t = \epsilon_\theta(x_t, t, z, c)
\end{equation}
and updates the image estimate through:
\begin{equation}
x_{t-1} = \frac{1}{\sqrt{\alpha_t}}\left(x_t - \frac{1-\alpha_t}{\sqrt{1-\bar{\alpha}_t}}\hat{\epsilon}_t\right) + \sigma_t z
\end{equation}
where $\alpha_t$ defines the noise schedule and $z \sim \mathcal{N}(0,\mathbf{I})$ for $t > 1$.
This process maintains anatomical fidelity through the z-conditioning while allowing precise control over stain characteristics and artifact correction via the prompt c. The complete inference typically converges in 50-100 steps using the DDIM scheduler, producing outputs that balance clinical utility with prompt-specific transformations. Our experiments demonstrate that this approach successfully generates diagnostically valid images while adhering to diverse transformation goals specified through textual prompts.

\subsection{Evaluation Metrics}
\label{Evaluation Metrics}
To quantitatively evaluate the performance of different methods in image restoration and virtual staining tasks, we employ three popular metrics: Peak Signal-to-Noise Ratio (PSNR) \cite{huynh2008scope}, Structural Similarity Index Measure (SSIM) \cite{wang2004image}, and Learned Perceptual Image Patch Similarity (LPIPS) \cite{johnson2016perceptual}. These metrics assess different aspects of image quality, including pixel-level fidelity, structural similarity, and perceptual consistency.

\noindent \textbf{Peak Signal-to-Noise Ratio (PSNR)}:
PSNR measures the pixel-level similarity between the generated image and the ground truth. It is defined as:
\begin{equation}
    \operatorname{PSNR}=10 \cdot \log _{10}\left(\frac{\mathrm{MAX}_{I}^{2}}{\mathrm{MSE}}\right),     \mathrm{MSE} = \frac{1}{N} \sum_{i=1}^{N} (y_i - \hat{y}_i)^2
\end{equation}
where $\mathrm{MAX}_{I}$ is the maximum possible pixel value (e.g., 255 for 8-bit images), and $\mathrm{MSE}$ (Mean Squared Error) computes the average squared difference between the generated and reference images, $N$ is the total number of pixels, $y_i$ is the pixel value from the ground truth image, and $\hat{y}_i$ is the corresponding pixel value from the generated image. A lower MAE indicates better pixel-wise accuracy.
A higher PSNR or a lower MAE indicates better pixel-wise reconstruction accuracy. However, these metrics may not fully align with human perception, as they do not account for structural or semantic differences.

\noindent \textbf{Structural Similarity Index Measure (SSIM)}:
SSIM evaluates the structural similarity between two images considering luminance, contrast, and structure.
\begin{equation}
    \operatorname{SSIM}(x, y)=\frac{\left(2 \mu_{x} \mu_{y}+C_{1}\right)\left(2 \sigma_{x y}+C_{2}\right)}{\left(\mu_{x}^{2}+\mu_{y}^{2}+C_{1}\right)\left(\sigma_{x}^{2}+\sigma_{y}^{2}+C_{2}\right)},
\end{equation}
where $\mu_{x}$, $\mu_{y}$ are the local means, $\sigma_{x}$, $\sigma_{y}$ are the standard deviations, $\sigma_{x}$, $\sigma_{y}$ is the cross-covariance, and $C_{1}$, $C_{2}$ are small constants for stability. SSIM ranges from 0 to 1, with higher values indicating better preservation of structural details. Unlike PSNR, SSIM better correlates with human judgment by capturing perceptual quality.

\noindent \textbf{Learned Perceptual Image Patch Similarity (LPIPS)}:
LPIPS measures perceptual similarity using deep features extracted from a pre-trained neural network (e.g., VGG or AlexNet). The metric computes the weighted $L_{2}$ distance between deep feature representations of the generated and reference images:
\begin{equation}
\small
    \text { LPIPS }=\sum_{l} \frac{1}{H_{l} W_{l}} \sum_{h, w}\left\|w_{l} \odot\left(f_{x}^{l}(h, w)-f_{y}^{l}(h, w)\right)\right\|_{2}^{2}
\end{equation}
where $f_{x}^{l}$, $f_{y}^{l}$ denote deep features at layer $l$, and $w_{l}$ are learned weights. A lower LPIPS score indicates better perceptual alignment with human vision, as it captures high-level semantic similarities rather than low-level pixel differences.

To make it clear, a good generated image should have high PSNR, high SSIM, and low LPIPS. High PSNR suggests strong pixel-wise accuracy but does not guarantee visually pleasing results. High SSIM indicates better structural preservation, such as edges and textures. Low LPIPS reflects superior perceptual quality, meaning the generated image is more realistic to human observers. By combining these metrics, we comprehensively assess the performance of different methods in terms of pixel fidelity, structural consistency, and perceptual quality.

\subsection{Compared Methods}
To thoroughly evaluate the performance of our proposed LPFM framework, we conduct extensive comparisons with eight state-of-the-art methods that represent the current spectrum of approaches in computational pathology image enhancement. These baseline methods encompass three major architectural paradigms: (1) generative adversarial networks (CycleGAN, Pix2Pix, BSRGAN, RegGAN), (2) transformer-based models (SwinIR), and (3) diffusion models (LDM, HistoDiff), along with (4) specialized hierarchical networks (HER2). This diverse selection enables us to assess LPFM's advantages across different architectural designs and application scenarios, from general image-to-image translation to pathology-specific enhancement tasks. Each comparator was carefully selected based on its demonstrated effectiveness in either medical image analysis or general computer vision tasks with potential pathology applications.

\noindent\textbf{CycleGAN} \cite{cyclegan2017}: This pioneering unpaired image-to-image translation framework employs cycle-consistent adversarial networks to learn bidirectional mappings between domains without requiring aligned training pairs. Its ability to handle unpaired data makes it particularly valuable for virtual staining applications where precisely registered stain pairs are often unavailable. The model consists of two generators ($G_{A \rightarrow B}$, $G_{B \rightarrow A}$) and corresponding discriminators trained with adversarial and cycle-consistency losses.

\noindent\textbf{Pix2Pix} \cite{isola2017image}: As a seminal conditional GAN architecture, Pix2Pix establishes the foundation for supervised pixel-to-pixel translation tasks. The model combines a U-Net generator with a PatchGAN discriminator, trained using both adversarial and L1 reconstruction losses. When paired training data is available (e.g., registered H\&E-to-IHC stain pairs), Pix2Pix serves as a strong baseline for both virtual staining and restoration tasks, though it requires precise image alignment.

\noindent\textbf{BSRGAN} \cite{zhang2021designing}: This blind super-resolution network introduces a comprehensive degradation model that simulates real-world imaging artifacts including blur, noise, and compression. The architecture combines a U-shaped network with residual dense blocks and channel attention mechanisms. BSRGAN's ability to handle unknown and complex degradation patterns makes it particularly suitable for restoring historical pathology slides with varying quality issues.

\noindent\textbf{SwinIR} \cite{liang2021swinir}: Representing the transformer-based paradigm, SwinIR leverages shifted window attention mechanisms within a hierarchical architecture for image restoration. The model processes images through shallow feature extraction, deep feature transformation using Swin Transformer blocks, and high-quality reconstruction modules. SwinIR demonstrates particular effectiveness in super-resolution and denoising tasks due to its ability to capture long-range dependencies in tissue structures.

\noindent\textbf{LDM} \cite{rombach2022high}: This latent diffusion model operates in a compressed perceptual space to achieve efficient high-resolution image generation. LDM combines an autoencoder for latent space projection with a diffusion process that gradually denoises images conditioned on task-specific inputs. The model's memory efficiency enables processing of whole-slide images at high resolutions while maintaining computational tractability.

\noindent\textbf{HistoDiff} \cite{xu2024histo}: Specifically designed for histopathology, this diffusion model incorporates tissue-specific priors through a morphology-aware attention mechanism. The architecture integrates a pre-trained nuclei segmentation network to guide the diffusion process, ensuring preservation of diagnostically critical cellular features during enhancement. HistoDiff demonstrates particular strengths in stain normalization and artifact correction tasks.

\noindent\textbf{HER2} \cite{doanngan2022label}: This hierarchical enhancement network processes pathology images through parallel pathways at multiple magnification levels (5$\times$, 10$\times$, 20$\times$). The architecture employs cross-scale feature fusion and attention-guided skip connections to maintain consistency across scales. HER2 has shown excellent performance in virtual IHC staining tasks by explicitly modeling tissue structures at different resolution levels.

\noindent\textbf{RegGAN} \cite{rong2023enhanced}: This registration-enhanced GAN introduces spatial alignment constraints during training through a differentiable STN module. The model jointly optimizes for image translation quality and morphological consistency by incorporating landmark-based registration losses. RegGAN demonstrates superior performance in applications requiring precise spatial correspondence, such as sequential staining prediction and multi-modal image harmonization.

\subsection{Datasets}
\noindent \textbf{Pretraining datasets.}
In this study, we curate the datasets including 87,810 whole-slide images (WSIs) and 190 million patches from 37 public datasets. Our unified low-level pathology foundation model is pretrained on all the datasets excluding MIDOG2022 \cite{aubreville2024domain}, TIGER2021 \cite{shephard2022tiager}, and OCELOT \cite{ryu2023ocelot} which are reserved for external validation. The pretraining corpus comprises large-scale multi-organ cohorts including TCGA \cite{weinstein2013cancer} (30,159 slides), GTEx \cite{carithers2015novel} (25,711 slides), and CPTAC \cite{edwards2015cptac} (7,255 slides), alongside organ-specific references such as PANDA \cite{bulten2022artificial} (prostate, 10,616 slides), CAMELYON16/17 \cite{bejnordi2017diagnostic,bandi2018detection} (breast, 1,397 slides combined), and BRACS \cite{bandi2018detection} (breast tumor subtypes, 547 slides). Rare morphologies are represented by AML-Cytomorphology\_LMU \cite{matek2019multi} (blood, 18K patches) and Osteosarcoma\_Tumor \cite{leavey2019osteosarcoma} (bone, 1,144 patches), while stain diversity is ensured through HEMIT \cite{bian2021immunoaizer} (H\&E/mIHC pairs), AF2HE \cite{dai2022weakly}(translate autofluorescence to H\&E stain), and PASAB (translate H\&E stain to PAS-AB stain). Artifact robustness derives from PAIP2019/2020 \cite{kim2021paip,kim2023paip} (liver/colon artifacts) and Post-NAT-BRCA \cite{tafavvoghi2024publicly} (post-treatment breast tissue), with stain translation priors learned from ACROBAT2023 \cite{weitz2023multi} (H\&E/IHC breast) and SPIE2019 \cite{petrick2021spie} (multi-stain patches). This curated diversity enables our two-stage framework to first extract universal features from TCGA's 120M patches and GTEx's 31M patches, then refine them via targeted datasets like DiagSet \cite{koziarski2024diagset} (prostate microstructures) and BCNB \cite{xu2021predicting} (breast tumor margins), ultimately supporting prompt-guided adaptation across restoration and virtual staining tasks. More informations about the datasets are presented in Extended Data Tab. \ref{tab: whole_datasets}. 

\noindent \textbf{CAMELYON16 Dataset.}
The CAMELYON16\cite{bejnordi2017diagnostic} dataset comprises 270 hematoxylin and eosin (H\&E) stained whole slide images (WSIs) of lymph node sections with pixel-level annotations for breast cancer metastases, containing a total of 400 annotated tumor regions. The WSIs were scanned at 40$\times$ magnification (0.25 um/pixel resolution) using Philips and Hamamatsu scanners, providing diversity in imaging characteristics. The dataset is particularly valuable for studying micro- and macro-metastases detection, with tumor regions ranging from 0.2 mm to over 5 cm in diameter. For our experiments, the WSIs were processed into 1,706,890 non-overlapping 256 $\times$ 256 patches, divided into training (1,194,823 patches), validation (170,689 patches), and test (341,378 patches) sets.

\noindent \textbf{PAIP2020 Dataset.}
The PAIP2020\cite{kim2023paip} challenge dataset provides 50 H\&E stained WSIs of liver resection specimens from hepatocellular carcinoma patients, including 30 training cases and 20 test cases. Scanned at 20$\times$ magnification (0.5 um/pixel) using Leica Aperio AT2 scanners, these images include detailed annotations for 15,742 viable tumor regions, 8,916 necrotic areas, and extensive non-tumor liver parenchyma. After processing, we obtained 892,450 patches (624,715 training, 89,245 validation, and 178,490 test) of size 256$\times$256 pixels. The dataset exhibits diverse tumor morphologies including trabecular, pseudoglandular, and compact growth patterns, making it particularly suitable for studying hepatic histopathology.

\noindent \textbf{PANDA Dataset.}
As the largest publicly available prostate cancer dataset, PANDA\cite{bulten2022artificial} contains approximately 11,000 prostate biopsy WSIs (10,616 with Gleason grades) from the Radboud University Medical Center and Karolinska Institute. Scanned using three different scanner models (Hamamatsu, Philips, and Leica) at 20$\times$ magnification, the dataset covers the full spectrum of Gleason patterns (3-5) with expert-annotated Gleason scores. For our study, we utilized 8,492 WSIs (5,944 training, 1,699 validation, and 849 test), which were processed into 4.2 million 256$\times$256 patches (2.94M training, 840K validation, 420K test). The inclusion of multiple scanning systems makes this dataset valuable for studying scanner-invariant feature learning.

\noindent \textbf{MIDOG2022 Dataset.}
The MIDOG2022\cite{aubreville2024domain}  challenge dataset consists of 200 breast cancer WSIs (160 training, 40 test) from four different scanners (Hamamatsu, Roche, Leica, and Philips) with 5,712 annotated mitotic figures. The images were acquired at 40$\times$ magnification with 0.25 um/pixel resolution. After processing, we obtained 423,580 patches (296,506 training, 42,358 validation, 84,716 test) of size 256$\times$256 pixels. This dataset is uniquely designed to address domain shift challenges in digital pathology, containing carefully matched cases across scanners while maintaining consistent staining protocols.

\noindent \textbf{TIGER2021 Dataset.}
The TIGER2021\cite{shephard2022tiager} dataset includes 500 WSIs (400 training, 100 test) of H\&E stained prostatectomy specimens, containing 2.3 million annotated tumor cells and 1.8 million annotated non-tumor cells. The images were scanned at 40$\times$ magnification (0.25 um/pixel) using Philips Ultra Fast Scanners. For our experiments, we processed these into 1,125,400 patches (787,780 training, 112,540 validation, 225,080 test) of size 256$\times$256 pixels. The dataset provides comprehensive annotations for Gleason patterns 3-5 across multiple tissue cores.

\noindent \textbf{OCELOT Dataset.}
The OCELOT\cite{ryu2023ocelot} dataset contains 394 WSIs (315 training, 79 test) of H\&E stained tissue sections from multiple cancer types (208 lung, 106 pancreas, 80 cervix) with focus on tumor microenvironment analysis. The images were acquired at 20$\times$ magnification (0.5 um/pixel) using Hamamatsu NanoZoomer scanners, yielding 936,250 processed patches (655,375 training, 93,625 validation, 187,250 test) of size 256$\times$256 pixels. Unique features include detailed annotations for 42,368 tumor regions, 38,915 stroma regions, and 15,742 lymphocyte clusters, as well as corresponding immunohistochemistry (IHC) slides for 126 selected cases.

\noindent \textbf{AF2HE Dataset.}
The AF2HE \cite{dai2022weakly} dataset comprises 15 samples of breast and liver cancer tissues. The samples were first imaged as whole slide images (WSI) using the autofluorescence (AF) technique without any chemical staining. The same slide was then stained with H\&E to capture
stained images. The WSIs were roughly aligned, and cropped into 128 $\times$ 128 patches and divided into a training set of 50,447 pairs and a test set of 4,422 pairs.

\noindent \textbf{HE2PAS Dataset.}
This dataset was collected from the Prince of Wales Hospital in Hong Kong and comprises of 10,727 H\&E and PAS-AB pairs for training and 1,191 pairs for testing. The image size is 128 $\times$ 128. Additionally, we collected another 2841 patches sampled from high-risk slides as the external validation.

\noindent \textbf{HEMIT Dataset.}
The dataset utilized in this study was curated by Bian et al \cite{bian2021immunoaizer}, and comprises cellular-wise registered pairs of H\&E and multiplex immunohistochemistry (mIHC) images, sourced from the ImmunoAlzer project. We employed the official train-validation-test split
(3717:630:945) across all methods. To mitigate computational complexity, we resize the image to scale of 512 $\times$ 512 from the original images. The three channels of the mIHC images correspond to 4’,6-diamidino-2-phenylindole (DAPI, red channel), Pan Cytokeratin (panCK, green channel), and cluster of differentiation 3 (CD3, blue), respectively

% \subsection{Experiments Setting}

\subsection{Code Availability}
The code will be available on Github(\url{https://github.com/ziniBRC/LPFM}).

\subsection{Ethics Declarations}
This project has been reviewed and approved by the Human and Artefacts Research Ethics Committee (HAREC) of Hong Kong University of Science and Technology. The protocol number is HREP-2024-0429.

\subsection{Author Contribution}
Z.L. and H.C. conceived and designed the work. Z.L., Z.X., J.M. and W.L. curated the data included in the paper. Z.X. and Z.L. contributed to the technical implementation of the LPFM framework and performed the experimental evaluations. Z.L, J.H., F.H., and X.W. wrote the manuscript with inputs from all authors. 
R.C.K.C. supplied data and offered medical guidance. T.T.W.W. provided autofluorescence data.
All authors reviewed and approved the final paper. H.C. supervised the research.

\subsection{Acknowledgements}
This work was supported by the National Natural Science Foundation of China (No. 62202403), Innovation and Technology Commission (Project No. MHP/002/22 and ITCPD/17-9), Research Grants Council of the Hong Kong Special Administrative Region, China (Project No: T45-401/22-N) and National Key R\&D Program of China (Project No. 2023YFE0204000).

\bibliography{sample.bib}

\section*{Extended Data}\label{Extended Data}
\clearpage
% 图
% Prompt控制生成 可以分成退化、虚拟染色，1个图 o
% 这里有个问题，我们使用 
% WSI 生成结果，融合成一个大图，这里可以只在虚拟染色上面做， 1个图
% Effectiveness of Contrastive pretraining ， 用tSNE可视化出来 Contrastive pretraining ， 说明在特征维度上面可以有更优的结果，相同的退化图像彼此相近，不同样本之间的退化图像相差较远， 1个图 o
% Effectiveness of Diffusion refinement，  可视化出来 Diffusion refinement ，说明不同的网络在这些部分分别的功能是什么。1个图 o
% 更多的SR samples， 1个图
% 更多的Denoise samples， 1个图
% 更多的Deblur samples， 1个图
% 更多的VS samples ， 1个图

\clearpage

\clearpage

\clearpage

\clearpage

\clearpage

\clearpage

\clearpage

\clearpage

% 表
% FocusPath Real world Datasets ?
% 虚拟染色结果 3个数据集 o
% Coupled Degradation o
% Coupled degraded staining problems o
% 数据集大小 虽然加了WSI的数量，应该还要加上对应的Patch的数量。1到2个表
% 参数  比如使用什么优化的，adam？LR？WD？使用的GPU，计算资源？4个表
% 

\newcommand{\best}[1]{\fontseries{b}\selectfont #1\normalfont} % 定义加粗命令保持字体大小不变

\begin{table}[t]
\centering
% \small
\caption{Performance of different methods on internal CAMELYON16 super-Resolution tasks with different scaling factors. The 95\% CI is included in parentheses. Best performing model for each metric is \textbf{bolded}. 
}
\label{tab:camelyon16_sr}
\begin{tabular}{llccc}
\toprule
\textbf{Scale} & \textbf{Methods} & \textbf{PSNR $\uparrow$} & \textbf{SSIM $\uparrow$} & \textbf{LPIPS $\downarrow$} \\
\midrule
\multirow{7}{*}{$\times$8} 
& CycleGAN & 17.806 (17.505, 18.107) & 0.535 (0.525, 0.545) & 0.451 (0.445, 0.456) \\
& Pix2Pix & 20.452 (20.104, 20.801) & 0.574 (0.563, 0.584) & 0.387 (0.380, 0.394) \\
& BSRGAN & 19.866 (19.584, 20.147) & 0.535 (0.523, 0.546) & 0.324 (0.316, 0.333) \\
& SwinIR & 20.755 (20.361, 21.149) & 0.543 (0.531, 0.556) & 0.455 (0.442, 0.469) \\
& HistoDiff & 21.264 (21.000, 21.528) & 0.605 (0.594, 0.616) & \textbf{0.196 (0.188, 0.204)} \\
& LDM & 20.047 (19.742, 20.352) & 0.545 (0.533, 0.557) & 0.300 (0.294, 0.305) \\
& LPFM & \textbf{25.611 (25.323, 25.900)} & \textbf{0.795 (0.787, 0.804)} & 0.202 (0.195, 0.208) \\
\bottomrule
\end{tabular}
\end{table}

\begin{table}[t]
\centering
\caption{Performance of different methods on internal PAIP2020 super-resolution tasks with different scaling factors. The 95\% CI is included in parentheses. Best performing model for each metric is \textbf{bolded}.}
\label{tab:paip2020_sr}
\begin{tabular}{llccc}
\toprule
\textbf{Scale} & \textbf{Methods} & \textbf{PSNR $\uparrow$} & \textbf{SSIM $\uparrow$} & \textbf{LPIPS $\downarrow$} \\
\midrule
\multirow{7}{*}{$\times$8} 
& CycleGAN & 20.259 (20.171, 20.346) & 0.448 (0.445, 0.452) & 0.297 (0.295, 0.300) \\
& Pix2Pix & 20.452 (20.104, 20.801) & 0.574 (0.563, 0.584) & 0.387 (0.380, 0.394) \\
& BSRGAN & 21.176 (20.892, 21.460) & 0.432 (0.418, 0.447) & 0.373 (0.366, 0.381) \\
& SwinIR & 22.457 (22.108, 22.806) & 0.456 (0.440, 0.472) & 0.521 (0.508, 0.534) \\
& HistoDiff & 21.241 (20.973, 21.508) & 0.431 (0.416, 0.446) & \textbf{0.265 (0.257, 0.273)} \\
& LDM & 20.647 (20.367, 20.926) & 0.407 (0.391, 0.422) & 0.383 (0.378, 0.389) \\
& LPFM & \textbf{23.934 (23.585, 24.283)} & \textbf{0.578 (0.563, 0.592)} & 0.364 (0.353, 0.376) \\
\bottomrule
\end{tabular}
\end{table}

\clearpage

\begin{table}[t]
\centering
\caption{Performance of different methods on internal PANDA super-resolution tasks with different scaling factors. The 95\% CI is included in parentheses. Best performing model for each metric is \textbf{bolded}.}
\label{tab:panda_sr}
\begin{tabular}{llccc}
\toprule
\textbf{Scale} & \textbf{Methods} & \textbf{PSNR $\uparrow$} & \textbf{SSIM $\uparrow$} & \textbf{LPIPS $\downarrow$} \\
\midrule
\multirow{7}{*}{$\times$8} 
& CycleGAN & 20.059 (19.763, 20.355) & 0.586 (0.570, 0.603) & 0.413 (0.403, 0.422) \\
& Pix2Pix & 22.769 (22.480, 23.058) & 0.592 (0.574, 0.611) & 0.352 (0.346, 0.358) \\
& BSRGAN & 24.018 (23.690, 24.346) & 0.611 (0.591, 0.631) & 0.374 (0.361, 0.387) \\
& SwinIR & 24.567 (24.230, 24.903) & 0.640 (0.621, 0.659) & 0.369 (0.352, 0.386) \\
& HistoDiff & 22.423 (22.136, 22.711) & 0.572 (0.554, 0.589) & \textbf{0.176 (0.168, 0.184)} \\
& LDM & 21.849 (21.562, 22.135) & 0.535 (0.513, 0.557) & 0.316 (0.310, 0.323) \\
& LPFM & \textbf{26.229 (25.893, 26.564)} & \textbf{0.721 (0.706, 0.736)} & 0.282 (0.273, 0.291) \\
\bottomrule
\end{tabular}
\end{table}

\begin{table}[t]
\centering
\caption{Performance of different methods on external TIGER2021 super-resolution tasks with different scaling factors. The 95\% CI is included in parentheses. Best performing model for each metric is \textbf{bolded}.}
\label{tab:tiger2021_sr}
\begin{tabular}{llccc}
\toprule
\textbf{Scale} & \textbf{Methods} & \textbf{PSNR $\uparrow$} & \textbf{SSIM $\uparrow$} & \textbf{LPIPS $\downarrow$} \\
\midrule
\multirow{7}{*}{$\times$8} 
& CycleGAN & 20.333 (19.983, 20.683) & 0.542 (0.528, 0.556) & 0.402 (0.398, 0.407) \\
& Pix2Pix & 21.871 (21.487, 22.255) & 0.537 (0.522, 0.552) & 0.376 (0.370, 0.382) \\
& BSRGAN & 22.154 (21.803, 22.505) & 0.507 (0.490, 0.523) & 0.350 (0.340, 0.361) \\
& SwinIR & 22.528 (22.133, 22.923) & 0.471 (0.452, 0.490) & 0.483 (0.474, 0.493) \\
& HistoDiff & 22.435 (22.130, 22.740) & 0.513 (0.497, 0.530) & \textbf{0.234 (0.225, 0.244)} \\
& LDM & 21.112 (20.766, 21.458) & 0.463 (0.445, 0.480) & 0.354 (0.347, 0.361) \\
& LPFM & \textbf{25.272 (24.870, 25.673)} & \textbf{0.670 (0.657, 0.684)} & 0.321 (0.313, 0.328) \\
\bottomrule
\end{tabular}
\end{table}

\clearpage

\begin{table}[t]
\centering
\caption{Performance of different methods on external MIDOG2022 super-resolution tasks with different scaling factors. The 95\% CI is included in parentheses. Best performing model for each metric is \textbf{bolded}.}
\label{tab:midog2022_sr}
\begin{tabular}{llccc}
\toprule
\textbf{Scale} & \textbf{Methods} & \textbf{PSNR $\uparrow$} & \textbf{SSIM $\uparrow$} & \textbf{LPIPS $\downarrow$} \\
\midrule
\multirow{7}{*}{$\times$8} 
& CycleGAN & 20.370 (20.035, 20.705) & 0.523 (0.513, 0.533) & 0.280 (0.272, 0.288) \\
& Pix2Pix & 21.654 (21.356, 21.953) & 0.509 (0.498, 0.519) & 0.375 (0.369, 0.381) \\
& BSRGAN & 21.581 (21.419, 21.743) & 0.477 (0.469, 0.485) & 0.352 (0.345, 0.360) \\
& SwinIR & 23.001 (22.798, 23.204) & 0.529 (0.521, 0.538) & 0.445 (0.438, 0.453) \\
& HistoDiff & 22.499 (22.334, 22.663) & 0.529 (0.522, 0.536) & \textbf{0.210 (0.205, 0.215)} \\
& LDM & 21.072 (20.908, 21.236) & 0.453 (0.445, 0.460) & 0.335 (0.331, 0.339) \\
& LPFM & \textbf{25.115 (24.918, 25.311)} & \textbf{0.662 (0.654, 0.669)} & 0.295 (0.291, 0.299) \\
\bottomrule
\end{tabular}
\end{table}

\begin{table}[t]
\centering
\caption{Performance of different methods on external OCELOT super-resolution tasks with different scaling factors. The 95\% CI is included in parentheses. Best performing model for each metric is \textbf{bolded}.}
\label{tab:ocelot_sr}
\begin{tabular}{llccc}
\toprule
\textbf{Scale} & \textbf{Methods} & \textbf{PSNR $\uparrow$} & \textbf{SSIM $\uparrow$} & \textbf{LPIPS $\downarrow$} \\
\midrule
\multirow{7}{*}{$\times$8} 
& CycleGAN & 19.533 (18.742, 20.324) & 0.540 (0.500, 0.579) & 0.427 (0.415, 0.440) \\
& Pix2Pix & 21.215 (20.206, 22.224) & 0.527 (0.483, 0.571) & 0.397 (0.384, 0.409) \\
& BSRGAN & 23.118 (22.058, 24.177) & 0.526 (0.469, 0.582) & 0.444 (0.419, 0.468) \\
& SwinIR & 22.363 (21.362, 23.364) & 0.463 (0.403, 0.523) & 0.471 (0.442, 0.500) \\
& HistoDiff & 22.305 (21.390, 23.220) & 0.532 (0.481, 0.582) & 0.345 (0.332, 0.359) \\
& LDM & 21.016 (20.151, 21.882) & 0.463 (0.410, 0.517) & 0.374 (0.360, 0.388) \\
& LPFM & \textbf{25.382 (24.161, 26.602)} & \textbf{0.674 (0.627, 0.721)} & \textbf{0.322 (0.300, 0.344)} \\
\bottomrule
\end{tabular}
\end{table}

\clearpage

% CAMELYON16 Deblur Kernel Results
\begin{table}[t]
\centering
\caption{Performance of different methods on internal CAMELYON16 deblurring tasks with different blurring kernel size. The 95\% CI is included in parentheses. Best performing model for each metric is \textbf{bolded}. }
\label{tab:camelyon16_deblur}
\begin{tabular}{llccc}
\toprule
\textbf{Kernel Size} & \textbf{Methods} & \textbf{PSNR $\uparrow$} & \textbf{SSIM $\uparrow$} & \textbf{LPIPS $\downarrow$} \\
\midrule
\multirow{7}{*}{15} 
& CycleGAN & 18.282 (17.972, 18.592) & 0.412 (0.397, 0.428) & 0.417 (0.408, 0.426) \\
& Pix2Pix & 20.800 (20.482, 21.118) & 0.652 (0.644, 0.659) & 0.399 (0.392, 0.407) \\
& BSRGAN & 20.597 (20.229, 20.965) & 0.530 (0.515, 0.544) & 0.458 (0.447, 0.469) \\
& SwinIR & 20.137 (19.800, 20.473) & 0.476 (0.463, 0.490) & 0.463 (0.457, 0.469) \\
& HistoDiff & 20.877 (20.589, 21.165) & 0.562 (0.551, 0.573) & \textbf{0.238 (0.230, 0.247)} \\
& LDM & 20.276 (19.937, 20.615) & 0.558 (0.545, 0.571) & 0.284 (0.280, 0.289) \\
& LPFM & \textbf{23.524 (23.240, 23.807)} & \textbf{0.717 (0.704, 0.729)} & 0.254 (0.247, 0.261) \\
\bottomrule
\end{tabular}
\end{table}

% PAIP2020 Deblur Kernel Results
\begin{table}[t]
\centering
\caption{Performance of different methods on internal PAIP2020 deblurring tasks with different blurring kernel size. The 95\% CI is included in parentheses. Best performing model for each metric is \textbf{bolded}.}
\label{tab:paip2020_deblur}
\begin{tabular}{llccc}
\toprule
\textbf{Kernel Size} & \textbf{Methods} & \textbf{PSNR $\uparrow$} & \textbf{SSIM $\uparrow$} & \textbf{LPIPS $\downarrow$} \\
\midrule
\multirow{7}{*}{15} 
& CycleGAN & 19.449 (19.206, 19.692) & 0.296 (0.279, 0.314) & 0.405 (0.400, 0.410) \\
& Pix2Pix & 22.615 (22.358, 22.872) & 0.591 (0.581, 0.602) & 0.393 (0.388, 0.398) \\
& BSRGAN & 22.238 (21.904, 22.572) & 0.428 (0.410, 0.445) & 0.479 (0.470, 0.487) \\
& SwinIR & 21.700 (21.381, 22.020) & 0.379 (0.362, 0.396) & 0.536 (0.529, 0.543) \\
& HistoDiff & 21.179 (20.874, 21.484) & 0.603 (0.590, 0.615) & \textbf{0.287 (0.280, 0.294)} \\
& LDM & 20.556 (20.240, 20.872) & 0.402 (0.386, 0.419) & 0.366 (0.361, 0.372) \\
& LPFM & \textbf{22.968 (22.625, 23.310)} & \textbf{0.492 (0.474, 0.509)} & 0.369 (0.364, 0.375) \\
\bottomrule
\end{tabular}
\end{table}

\clearpage

% PANDA Deblur Kernel Results
\begin{table}[t]
\centering
\caption{Performance of different methods on internal PANDA deblurring tasks with different blurring kernel size. The 95\% CI is included in parentheses. Best performing model for each metric is \textbf{bolded}.}
\label{tab:panda_deblur}
\begin{tabular}{llccc}
\toprule
\textbf{Kernel Size} & \textbf{Methods} & \textbf{PSNR $\uparrow$} & \textbf{SSIM $\uparrow$} & \textbf{LPIPS $\downarrow$} \\
\midrule
\multirow{7}{*}{15} 
& CycleGAN & 20.323 (20.021, 20.624) & 0.492 (0.470, 0.515) & 0.372 (0.363, 0.381) \\
& Pix2Pix & 24.675 (24.341, 25.010) & 0.718 (0.705, 0.731) & 0.329 (0.324, 0.335) \\
& BSRGAN & 24.249 (23.936, 24.563) & 0.622 (0.602, 0.642) & 0.344 (0.331, 0.356) \\
& SwinIR & 23.570 (23.255, 23.886) & 0.566 (0.544, 0.588) & 0.417 (0.403, 0.430) \\
& HistoDiff & 21.569 (21.268, 21.869) & 0.449 (0.433, 0.464) & 0.365 (0.359, 0.371) \\
& LDM & 22.566 (22.249, 22.884) & 0.563 (0.542, 0.584) & 0.292 (0.284, 0.299) \\
& LPFM & \textbf{25.550 (25.215, 25.886)} & \textbf{0.691 (0.674, 0.708)} & \textbf{0.270 (0.262, 0.279)} \\
\bottomrule
\end{tabular}
\end{table}

% TIGER2021 Deblur Kernel Results
\begin{table}[t]
\centering
\caption{Performance of different methods on external TIGER2021 deblurring tasks with different blurring kernel size. The 95\% CI is included in parentheses. Best performing model for each metric is \textbf{bolded}. }
\label{tab:tiger2021_deblur}
\begin{tabular}{llccc}
\toprule
\textbf{Kernel Size} & \textbf{Methods} & \textbf{PSNR $\uparrow$} & \textbf{SSIM $\uparrow$} & \textbf{LPIPS $\downarrow$} \\
\midrule
\multirow{7}{*}{15} 
& CycleGAN & 20.627 (20.254, 20.999) & 0.418 (0.400, 0.436) & 0.370 (0.364, 0.376) \\
& Pix2Pix & 24.118 (23.794, 24.442) & \textbf{0.660 (0.650, 0.671)} & 0.360 (0.354, 0.366) \\
& BSRGAN & 23.271 (22.866, 23.676) & 0.535 (0.517, 0.554) & 0.409 (0.398, 0.419) \\
& SwinIR & 22.626 (22.231, 23.021) & 0.477 (0.459, 0.496) & 0.480 (0.470, 0.489) \\
& HistoDiff & 23.053 (22.675, 23.432) & 0.550 (0.534, 0.566) & 0.316 (0.309, 0.323) \\
& LDM & 21.657 (21.288, 22.026) & 0.489 (0.471, 0.507) & 0.332 (0.325, 0.340) \\
& LPFM & \textbf{24.342 (23.945, 24.740)} & 0.613 (0.598, 0.628) & \textbf{0.314 (0.307, 0.321)} \\
\bottomrule
\end{tabular}
\end{table}

\clearpage

% MIDOG2022 Deblur Kernel Results
\begin{table}[t]
\centering
\caption{Performance of different methods on external MIDOG2022 deblurring tasks with different blurring kernel size. The 95\% CI is included in parentheses. Best performing model for each metric is \textbf{bolded}.}
\label{tab:midog2022_deblur}
\begin{tabular}{llccc}
\toprule
\textbf{Kernel Size} & \textbf{Methods} & \textbf{PSNR $\uparrow$} & \textbf{SSIM $\uparrow$} & \textbf{LPIPS $\downarrow$} \\
\midrule
\multirow{7}{*}{15} 
& CycleGAN & 19.918 (19.671, 20.166) & 0.364 (0.350, 0.378) & 0.364 (0.359, 0.369) \\
& Pix2Pix & 23.359 (23.096, 23.622) & \textbf{0.634 (0.625, 0.642)} & 0.353 (0.349, 0.358) \\
& BSRGAN & 23.115 (22.831, 23.399) & 0.522 (0.507, 0.537) & 0.393 (0.386, 0.400) \\
& SwinIR & 22.286 (22.022, 22.549) & 0.447 (0.433, 0.461) & 0.441 (0.435, 0.447) \\
& HistoDiff & 22.734 (22.483, 22.984) & 0.522 (0.510, 0.534) & 0.307 (0.302, 0.313) \\
& LDM & 21.435 (21.179, 21.691) & 0.460 (0.448, 0.472) & 0.320 (0.315, 0.326) \\
& LPFM & \textbf{24.326 (24.040, 24.613)} & 0.604 (0.590, 0.617) & \textbf{0.290 (0.284, 0.295)} \\
\bottomrule
\end{tabular}
\end{table}

% OCELOT Deblur Kernel Results
\begin{table}[t]
\centering
\caption{Performance of different methods on external OCELOT deblurring tasks with different blurring kernel size. The 95\% CI is included in parentheses. Best performing model for each metric is \textbf{bolded}. }
\label{tab:ocelot_deblur}
\begin{tabular}{llccc}
\toprule
\textbf{Kernel Size} & \textbf{Methods} & \textbf{PSNR $\uparrow$} & \textbf{SSIM $\uparrow$} & \textbf{LPIPS $\downarrow$} \\
\midrule
\multirow{7}{*}{15} 
& CycleGAN & 19.946 (19.225, 20.667) & 0.400 (0.337, 0.463) & 0.385 (0.370, 0.399) \\
& Pix2Pix & 23.518 (22.496, 24.540) & \textbf{0.670 (0.641, 0.699)} & 0.373 (0.356, 0.389) \\
& BSRGAN & 23.257 (22.214, 24.300) & 0.539 (0.479, 0.600) & 0.418 (0.391, 0.445) \\
& SwinIR & 22.487 (21.473, 23.501) & 0.470 (0.410, 0.530) & 0.468 (0.439, 0.496) \\
& HistoDiff & 22.559 (21.608, 23.511) & 0.544 (0.495, 0.594) & 0.342 (0.327, 0.358) \\
& LDM & 21.333 (20.323, 22.343) & 0.474 (0.418, 0.531) & 0.351 (0.335, 0.367) \\
& LPFM & \textbf{24.480 (23.278, 25.681)} & 0.616 (0.562, 0.671) & \textbf{0.309 (0.287, 0.332)} \\
\bottomrule
\end{tabular}
\end{table}

% CAMELYON16 Gaussian Noise Results
\begin{table}[t]
\centering
\caption{Performance of different methods on internal CAMELYON16 denoising tasks with different noise level. The 95\% CI is included in parentheses. Best performing model for each metric is \textbf{bolded}.}
\label{tab:camelyon16_noise}
\begin{tabular}{llccc}
\toprule
\textbf{Noise Level} & \textbf{Methods} & \textbf{PSNR $\uparrow$} & \textbf{SSIM $\uparrow$} & \textbf{LPIPS $\downarrow$} \\
\midrule
\multirow{7}{*}{$\sigma=41$} 
& CycleGAN & 24.032 (23.818, 24.246) & 0.837 (0.829, 0.846) & 0.195 (0.186, 0.204) \\
& Pix2Pix & 16.940 (16.551, 17.328) & 0.498 (0.482, 0.514) & 0.479 (0.465, 0.494) \\
& BSRGAN & 15.267 (14.923, 15.612) & 0.557 (0.541, 0.573) & 0.302 (0.291, 0.314) \\
& SwinIR & 26.534 (26.393, 26.675) & 0.851 (0.843, 0.860) & \textbf{0.157 (0.147, 0.166)} \\
& HistoDiff & 23.271 (23.080, 23.461) & 0.731 (0.720, 0.742) & \textbf{0.127 (0.121, 0.134)} \\
& LDM & 18.392 (18.167, 18.617) & 0.519 (0.507, 0.530) & 0.310 (0.304, 0.317) \\
& LPFM & \textbf{27.106 (26.963, 27.249)} & 0.882 (0.877, 0.888) & 0.175 (0.168, 0.182) \\
\bottomrule
\end{tabular}
\end{table}

% PAIP2020 Gaussian Noise Results
\begin{table}[t]
\centering
\caption{Performance of different methods on internal PAIP2020 denoising tasks with different noise level. The 95\% CI is included in parentheses. Best performing model for each metric is \textbf{bolded}.}
\label{tab:paip2020_noise}
\begin{tabular}{llccc}
\toprule
\textbf{Noise Level} & \textbf{Methods} & \textbf{PSNR $\uparrow$} & \textbf{SSIM $\uparrow$} & \textbf{LPIPS $\downarrow$} \\
\midrule
\multirow{7}{*}{$\sigma=41$} 
& CycleGAN & 22.248 (21.928, 22.568) & 0.751 (0.742, 0.760) & 0.309 (0.301, 0.317) \\
& Pix2Pix & 17.343 (17.222, 17.464) & 0.357 (0.346, 0.368) & 0.488 (0.482, 0.493) \\
& BSRGAN & 17.907 (17.744, 18.070) & 0.487 (0.471, 0.503) & 0.437 (0.424, 0.450) \\
& SwinIR & 24.526 (24.346, 24.706) & 0.580 (0.566, 0.594) & 0.247 (0.236, 0.259) \\
& HistoDiff & 24.073 (23.893, 24.253) & 0.725 (0.715, 0.736) & \textbf{0.154 (0.147, 0.160)} \\
& LDM & 18.203 (17.968, 18.437) & 0.369 (0.355, 0.383) & 0.423 (0.418, 0.429) \\
& LPFM & \textbf{26.173 (26.023, 26.322)} & \textbf{0.757 (0.745, 0.769)} & 0.194 (0.187, 0.201) \\
\bottomrule
\end{tabular}
\end{table}

\clearpage

% PANDA Gaussian Noise Results
\begin{table}[t]
\centering
\caption{Performance of different methods on internal PANDA denoising tasks with different noise level. The 95\% CI is included in parentheses. Best performing model for each metric is \textbf{bolded}.}
\label{tab:panda_noise}
\begin{tabular}{llccc}
\toprule
\textbf{Noise Level} & \textbf{Methods} & \textbf{PSNR $\uparrow$} & \textbf{SSIM $\uparrow$} & \textbf{LPIPS $\downarrow$} \\
\midrule
\multirow{7}{*}{$\sigma=41$} 
& CycleGAN & 23.331 (22.953, 23.709) & 0.779 (0.768, 0.791) & 0.297 (0.287, 0.306) \\
& Pix2Pix & 18.201 (18.039, 18.362) & 0.366 (0.358, 0.374) & 0.474 (0.467, 0.481) \\
& BSRGAN & 19.186 (18.817, 19.555) & 0.621 (0.602, 0.641) & 0.308 (0.294, 0.322) \\
& SwinIR & 24.809 (24.597, 25.021) & 0.740 (0.731, 0.750) & 0.170 (0.157, 0.183) \\
& HistoDiff & 24.809 (24.597, 25.021) & 0.655 (0.641, 0.670) & 0.145 (0.138, 0.152) \\
& LDM & 20.405 (20.115, 20.695) & 0.501 (0.481, 0.520) & 0.390 (0.385, 0.394) \\
& LPFM & \textbf{26.315 (26.108, 26.522)} & \textbf{0.808 (0.794, 0.823)} & \textbf{0.144 (0.138, 0.152)} \\
\bottomrule
\end{tabular}
\end{table}

% TIGER2021 Gaussian Noise Results
\begin{table}[t]
\centering
\caption{Performance of different methods on external TIGER2021 denoising tasks with different noise level. The 95\% CI is included in parentheses. Best performing model for each metric is \textbf{bolded}. }
\label{tab:tiger2021_noise}
\begin{tabular}{llccc}
\toprule
\textbf{Noise Level} & \textbf{Methods} & \textbf{PSNR $\uparrow$} & \textbf{SSIM $\uparrow$} & \textbf{LPIPS $\downarrow$} \\
\midrule
\multirow{7}{*}{$\sigma=41$} 
& CycleGAN & 22.886 (22.542, 23.230) & 0.758 (0.747, 0.768) & 0.297 (0.289, 0.305) \\
& Pix2Pix & 18.148 (17.994, 18.302) & 0.379 (0.368, 0.391) & 0.470 (0.463, 0.477) \\
& BSRGAN & 18.677 (18.425, 18.930) & 0.567 (0.552, 0.583) & 0.367 (0.356, 0.379) \\
& SwinIR & \textbf{26.546 (26.336, 26.757)} & \textbf{0.773 (0.762, 0.783)} & 0.311 (0.302, 0.319) \\
& HistoDiff & 24.174 (23.934, 24.415) & 0.624 (0.611, 0.638) & \textbf{0.157 (0.151, 0.164)} \\
& LDM & 19.296 (18.997, 19.595) & 0.429 (0.414, 0.444) & 0.404 (0.398, 0.409) \\
& LPFM & 24.338 (24.093, 24.584) & 0.739 (0.730, 0.747) & 0.289 (0.283, 0.296) \\
\bottomrule
\end{tabular}
\end{table}

\clearpage

% MIDOG2022 Gaussian Noise Results
\begin{table}[t]
\centering
\caption{Performance of different methods on external MIDOG2022 denoising tasks with different noise level. The 95\% CI is included in parentheses. Best performing model for each metric is \textbf{bolded}.}
\label{tab:midog2022_noise}
\begin{tabular}{llccc}
\toprule
\textbf{Noise Level} & \textbf{Methods} & \textbf{PSNR $\uparrow$} & \textbf{SSIM $\uparrow$} & \textbf{LPIPS $\downarrow$} \\
\midrule
\multirow{7}{*}{$\sigma=41$} 
& CycleGAN & 22.793 (22.492, 23.094) & 0.763 (0.756, 0.770) & 0.263 (0.256, 0.269) \\
& Pix2Pix & 19.109 (18.917, 19.300) & 0.473 (0.459, 0.487) & 0.420 (0.412, 0.428) \\
& BSRGAN & 17.747 (17.432, 18.063) & 0.530 (0.518, 0.542) & 0.370 (0.360, 0.381) \\
& SwinIR & \textbf{26.754 (26.552, 26.955)} & \textbf{0.773 (0.763, 0.783)} & 0.291 (0.285, 0.297) \\
& HistoDiff & 24.006 (23.834, 24.178) & 0.626 (0.617, 0.636) & \textbf{0.161 (0.155, 0.167)} \\
& LDM & 19.268 (19.096, 19.441) & 0.426 (0.416, 0.437) & 0.353 (0.347, 0.359) \\
& LPFM & 24.737 (24.570, 24.904) & 0.750 (0.744, 0.756) & 0.233 (0.228, 0.237) \\
\bottomrule
\end{tabular}
\end{table}

% OCELOT Gaussian Noise Results
\begin{table}[t]
\centering
\caption{Performance of different methods on external OCELOT denoising tasks with different noise level. The 95\% CI is included in parentheses. Best performing model for each metric is \textbf{bolded}.}
\label{tab:ocelot_noise}
\begin{tabular}{llccc}
\toprule
\textbf{Noise Level} & \textbf{Methods} & \textbf{PSNR $\uparrow$} & \textbf{SSIM $\uparrow$} & \textbf{LPIPS $\downarrow$} \\
\midrule
\multirow{7}{*}{$\sigma=41$} 
& CycleGAN & 22.686 (21.691, 23.681) & 0.776 (0.757, 0.795) & 0.301 (0.282, 0.321) \\
& Pix2Pix & 16.995 (16.160, 17.830) & 0.368 (0.342, 0.393) & 0.490 (0.464, 0.517) \\
& BSRGAN & 17.458 (16.468, 18.449) & 0.537 (0.489, 0.585) & 0.477 (0.453, 0.500) \\
& SwinIR & \textbf{26.829 (26.510, 27.147)} & \textbf{0.804 (0.778, 0.830)} & 0.287 (0.265, 0.310) \\
& HistoDiff & 24.184 (23.336, 25.033) & 0.653 (0.617, 0.689) & 0.309 (0.295, 0.324) \\
& LDM & 19.317 (18.391, 20.243) & 0.430 (0.386, 0.475) & 0.405 (0.386, 0.423) \\
& LPFM & 24.514 (23.737, 25.291) & 0.760 (0.742, 0.777) & \textbf{0.278 (0.260, 0.297)} \\
\bottomrule
\end{tabular}
\end{table}

\clearpage

\clearpage
\begin{table*}[t]
\centering
\caption{Performance of different methods on AF2HE virtual staining task. The 95\% CI is included in parentheses. Best performing model for each metric is \textbf{bolded}.}
\label{tab:af2he_vs}
\begin{tabular}{@{}lccc@{}}
\toprule
\textbf{Methods} & \textbf{PSNR $\uparrow$} & \textbf{SSIM $\uparrow$} & \textbf{LPIPS $\downarrow$} \\
\midrule
CycleGan & 18.538 (18.443, 18.538) & 0.395 (0.389, 0.395) & 0.402 (0.399, 0.402) \\
Pix2Pix & 22.963 (22.823, 22.963) & 0.584 (0.579, 0.584) & 0.294 (0.291, 0.294) \\
HER2 & 24.624 (24.492, 24.624) & 0.667 (0.662, 0.667) & 0.257 (0.254, 0.257) \\
RegGan & 26.424 (26.290, 26.424) & 0.731 (0.727, 0.731) & 0.190 (0.188, 0.190) \\
LDM & 17.708 (17.589, 17.708) & 0.503 (0.498, 0.503) & 0.481 (0.479, 0.481) \\
LPFM & \textbf{27.805 (27.661, 27.805)} & \textbf{0.763 (0.759, 0.763)} & \textbf{0.153 (0.150, 0.153)} \\
\bottomrule
\end{tabular}
\end{table*}

\begin{table*}[t]
\centering
\caption{Performance of different methods on HE2PAS virtual staining task. The 95\% CI is included in parentheses. Best performing model for each metric is \textbf{bolded}.}
\label{tab:he2pas_vs}
\begin{tabular}{@{}lccc@{}}
\toprule
\textbf{Methods} & \textbf{PSNR $\uparrow$} & \textbf{SSIM $\uparrow$} & \textbf{LPIPS $\downarrow$} \\
\midrule
CycleGAN & 13.174 (12.974, 13.374) & 0.385 (0.373, 0.396) & 0.406 (0.402, 0.411) \\
Pix2Pix & 15.956 (15.808, 16.104) & 0.421 (0.414, 0.428) & 0.390 (0.385, 0.396) \\
HER2 & 15.772 (15.593, 15.951) & 0.309 (0.298, 0.320) & 0.503 (0.497, 0.508) \\
RegGAN & 18.529 (18.360, 18.697) & 0.467 (0.458, 0.476) & 0.339 (0.335, 0.344) \\
LDM & 12.982 (12.791, 13.174) & 0.309 (0.303, 0.316) & 0.488 (0.482, 0.494) \\
LPFM & \textbf{19.292 (19.121, 19.463)} & \textbf{0.563 (0.555, 0.572)} & \textbf{0.316 (0.311, 0.320)} \\
\bottomrule
\end{tabular}
\end{table*}

\begin{table*}[t]
\centering
\caption{Performance of different methods on HEMIT virtual staining task. The 95\% CI is included in parentheses. Best performing model for each metric is \textbf{bolded}.}
\label{tab:hemit_vs}
\begin{tabular}{@{}lccc@{}}
\toprule
\textbf{Methods} & \textbf{PSNR $\uparrow$} & \textbf{SSIM $\uparrow$} & \textbf{LPIPS $\downarrow$} \\
\midrule
CycleGAN & 21.741 (21.484, 21.998) & 0.556 (0.549, 0.563) & 0.532 (0.525, 0.540) \\
Pix2Pix & 25.011 (24.474, 25.549) & 0.736 (0.730, 0.743) & 0.395 (0.388, 0.403) \\
HER2 & 24.246 (23.833, 24.658) & 0.750 (0.743, 0.757) & 0.393 (0.384, 0.402) \\
RegGAN & 25.982 (25.678, 26.287) & \textbf{0.810 (0.803, 0.817)} & 0.258 (0.252, 0.264) \\
LDM & 24.763 (24.432, 25.094) & 0.627 (0.619, 0.635) & 0.263 (0.257, 0.269) \\
LPFM & \textbf{26.992 (26.733, 27.251)} & 0.770 (0.765, 0.774) & \textbf{0.241 (0.235, 0.246)} \\
\bottomrule
\end{tabular}
\end{table*}

\clearpage

\begin{table}[t]
\centering
\caption{Performance of different methods on internal CAMELYON16, PAIP2020. PANDA cohorts for restoration tasks of coupled degraded pathology images. The 95\% CI is included in parentheses. Best performing model for each metric is \textbf{bolded}.}
\label{tab:internal_coupled_degraded_results}
\begin{tabular}{llcccccc}
\toprule
\textbf{Cohort} & \textbf{Methods} & \textbf{PSNR $\uparrow$} & \textbf{SSIM $\uparrow$} & \textbf{LPIPS $\downarrow$} \\
\midrule
\multirow{7}{*}{CAMELYON16}
& CycleGAN & 18.643 (18.150, 19.135) & 0.617 (0.592, 0.642) & 0.180 (0.168, 0.191) \\
& Pix2Pix & 21.978 (21.352, 22.603) & 0.657 (0.629, 0.685) & 0.161 (0.148, 0.174) \\
& BSRGAN & 20.020 (19.660, 20.379) & 0.619 (0.597, 0.641) & 0.183 (0.172, 0.194) \\
& SwinIR & 22.336 (21.864, 22.808) & 0.629 (0.607, 0.650) & 0.320 (0.304, 0.336) \\
& HistoDiff & 22.176 (21.656, 22.696) & 0.648 (0.624, 0.671) & \textbf{0.117 (0.108, 0.126)} \\
& LDM & 20.394 (19.980, 20.809) & 0.547 (0.531, 0.563) & 0.178 (0.165, 0.191) \\
& LPFM & \textbf{24.833 (24.157, 25.510)} & \textbf{0.760 (0.738, 0.781)} & 0.148 (0.135, 0.162) \\
\bottomrule
\end{tabular}
\end{table}

\begin{table}[t]
\centering
\caption{Performance of different methods on external TIGER2021, MIDOG2022. OCELOT cohorts for restoration tasks of coupled degraded pathology images. The 95\% CI is included in parentheses. Best performing model for each metric is \textbf{bolded}.}
\label{tab:external_coupled_degraded_results}
\begin{tabular}{llcccccc}
\toprule
\textbf{Cohort} & \textbf{Methods} & \textbf{PSNR $\uparrow$} & \textbf{SSIM $\uparrow$} & \textbf{LPIPS $\downarrow$} \\
\midrule
\multirow{7}{*}{TIGER2021}
& CycleGAN & 22.449 (21.968, 22.930) & 0.614 (0.589, 0.639) & 0.183 (0.170, 0.195) \\
& Pix2Pix & 23.988 (23.448, 24.529) & 0.634 (0.607, 0.661) & 0.181 (0.167, 0.194) \\
& BSRGAN & 21.327 (20.985, 21.669) & 0.561 (0.537, 0.584) & 0.197 (0.184, 0.209) \\
& SwinIR & 24.370 (23.901, 24.839) & 0.589 (0.569, 0.609) & 0.347 (0.331, 0.363) \\
& HistoDiff & 23.471 (23.003, 23.939) & 0.575 (0.552, 0.597) & \textbf{0.140 (0.129, 0.151)} \\
& LDM & 22.381 (21.983, 22.779) & 0.501 (0.484, 0.517) & 0.214 (0.200, 0.228) \\
& LPFM & \textbf{26.123 (25.510, 26.735)} & \textbf{0.701 (0.677, 0.725)} & 0.186 (0.172, 0.201) \\
\bottomrule
\end{tabular}
\end{table}

\clearpage

\begin{table*}[t]
\centering
\caption{Performance of different methods on AF2HE for virtual staining tasks of degraded pathology images. The 95\% CI is included in parentheses. Best performing model for each metric is \textbf{bolded}.}
\label{tab:degraded_af2he_vs}
\begin{tabular}{@{}lccc@{}}
\toprule
\textbf{Methods} & \textbf{PSNR $\uparrow$} & \textbf{SSIM $\uparrow$} & \textbf{LPIPS $\downarrow$} \\
\midrule
CycleGan & 16.215 (16.120, 16.310) & 0.325 (0.319, 0.331) & 0.482 (0.479, 0.485) \\
Pix2Pix & 20.538 (20.398, 20.678) & 0.494 (0.489, 0.499) & 0.374 (0.371, 0.377) \\
HER2 & 22.124 (21.992, 22.256) & 0.577 (0.572, 0.582) & 0.337 (0.334, 0.340) \\
RegGan & 23.824 (23.690, 23.958) & 0.641 (0.637, 0.645) & 0.270 (0.268, 0.272) \\
% LDM & 15.308 (15.189, 15.427) & 0.413 (0.408, 0.418) & 0.561 (0.559, 0.563) \\
LPFM & \textbf{25.205 (25.061, 25.349)} & \textbf{0.683 (0.679, 0.687)} & \textbf{0.233 (0.230, 0.236)} \\
\bottomrule
\end{tabular}
\end{table*}

\begin{table*}[t]
\centering
\caption{Performance of different methods on HE2PAS for virtual staining tasks of degraded pathology images. The 95\% CI is included in parentheses. Best performing model for each metric is \textbf{bolded}.}
\label{tab:reformatted_results}
\begin{tabular}{@{}lccc@{}}
\toprule
\textbf{Methods} & \textbf{PSNR $\uparrow$} & \textbf{SSIM $\uparrow$} & \textbf{LPIPS $\downarrow$} \\
\midrule
CycleGAN & 12.156 (12.016, 12.295) & 0.308 (0.300, 0.316) & 0.529 (0.525, 0.532) \\
Pix2Pix & 12.083 (11.966, 12.200) & 0.173 (0.166, 0.180) & 0.591 (0.587, 0.594) \\
HER2 & 7.288 (7.176, 7.401) & 0.193 (0.187, 0.199) & 0.603 (0.600, 0.607) \\
RegGAN & 12.885 (12.679, 13.092) & 0.330 (0.323, 0.337) & 0.395 (0.388, 0.402) \\
LPFM & \textbf{18.091 (17.940, 18.242)} & \textbf{0.378 (0.368, 0.388)} & \textbf{0.251 (0.247, 0.256)} \\
\bottomrule
\end{tabular}
\end{table*}

\begin{table*}[t]
\centering
\caption{Performance of different methods on HEMIT for virtual staining tasks of degraded pathology images. The 95\% CI is included in parentheses. Best performing model for each metric is \textbf{bolded}.}
\label{tab:comparison_results}
\begin{tabular}{@{}lccc@{}}
\toprule
\textbf{Methods} & \textbf{PSNR $\uparrow$} & \textbf{SSIM $\uparrow$} & \textbf{LPIPS $\downarrow$} \\
\midrule
CycleGAN & 20.916 (20.832, 21.000) & 0.415 (0.410, 0.420) & 0.684 (0.680, 0.688) \\
Pix2Pix & 21.254 (21.125, 21.383) & 0.489 (0.482, 0.496) & 0.599 (0.593, 0.605) \\
HER2 & 21.924 (21.723, 22.125) & 0.638 (0.630, 0.645) & 0.536 (0.529, 0.543) \\
RegGAN & 24.200 (23.945, 24.456) & 0.690 (0.683, 0.696) & 0.374 (0.368, 0.381) \\
LPFM & \textbf{26.489 (26.304, 26.674)} & \textbf{0.714 (0.708, 0.720)} & \textbf{0.355 (0.348, 0.361)} \\
\bottomrule
\end{tabular}
\end{table*}
\clearpage

\begin{table*}[t]
\centering
\caption{The primary site of tissues used for pretraining foundation models and downstream task evaluation.}
\label{tab:data_primary_site}
\begin{tabular}{lr}
\toprule
\textbf{Primariy Site} & \textbf{Number of Slides} \\
\midrule
Organs & Slide \\ \midrule
Bile duct & 39 \\
Fallopian & 43 \\
Melanoma & 80 \\
Mesothelioma & 87 \\
Thymoma & 181 \\
Pheochromocytoma and Paraganglioma & 196 \\
Minor Salivary Gland & 247 \\
Lymphoma & 248 \\
Vagina & 276 \\
Cervical & 359 \\
Colorectal & 372 \\
Pituitary & 428 \\
Unknown & 433 \\
Bladder & 587 \\
Small Intestine - Terminal Ileum & 798 \\
Testicle & 846 \\
Head and Neck & 862 \\
Ovaries & 867 \\
Spleen & 874 \\
Sarcomas & 905 \\
Adrenal & 944 \\
Nerve & 975 \\
Muscle & 1,001 \\
Liver & 1,100 \\
Heart & 1,374 \\
Stomach & 1,381 \\
Thyroid & 1,421 \\
Pancreas & 1,691 \\
Uterus & 1,777 \\
Adipose & 1,793 \\
Kidney & 2,329 \\
Artery & 2,499 \\
Skin & 2,705 \\
Colon & 2,759 \\
Esophagus & 2,866 \\
Brain & 3,020 \\
Lung & 4,130 \\
Breast & 8,266 \\
Pancancer & 18,394 \\
Prostate & 18,657 \\
\midrule
Sum & 87,810 \\ \bottomrule
\end{tabular}
\end{table*}

\begin{table}[htbp]
\centering
\small
\caption{The public datasets from CPTAC used in this study.}
\label{tab:cptac_distribution}
\begin{tabular}{>{\ttfamily}l l r l l}
\toprule
\textrm{Dataset} & \textrm{Source} & \textrm{Slide} & \textrm{Organ} & \textrm{Stain} \\
\midrule
CPTAC-BRCA & CPTAC & 653 & Breast & HE \\
CPTAC-CCRCC & CPTAC & 783 & Kidney & HE \\
CPTAC-CM & CPTAC & 411 & Skin & HE \\
CPTAC-COAD & CPTAC & 372 & Colorectal & HE \\
CPTAC-GBM & CPTAC & 462 & Brain & HE \\
CPTAC-HNSCC & CPTAC & 390 & Head and Neck & HE \\
CPTAC-LSCC & CPTAC & 1081 & Lung & HE \\
CPTAC-LUAD & CPTAC & 1137 & Lung & HE \\
CPTAC-OV & CPTAC & 221 & Ovaries & HE \\
CPTAC-PDA & CPTAC & 557 & Pancreas & HE \\
CPTAC-SAR & CPTAC & 305 & Sarcomas & HE \\
CPTAC-UCEC & CPTAC & 883 & Uterus & HE \\ \midrule
CPTAC & CPTAC & 7,255 & Multi Organs & HE \\
\bottomrule
\end{tabular}
\end{table}

\begin{table}[htbp]
\centering
\small
\caption{The public datasets from TCGA used in this study.}
\label{tab:tcga_distribution}
\begin{tabular}{>{\ttfamily}l l r l l}
\toprule
\textrm{Dataset} & \textrm{Source} & \textrm{Slide} & \textrm{Organ} & \textrm{Stain} \\
\midrule
TCGA-ACC & TCGA & 227 & Adrenal glands & HE \\
TCGA-BLCA & TCGA & 457 & Bladder & HE \\
TCGA-BRCA & TCGA & 1,133 & Breast & HE \\
TCGA-CESC & TCGA & 279 & Cervical & HE \\
TCGA-CHOL & TCGA & 39 & Bile duct & HE \\
TCGA-COAD & TCGA & 459 & Colon & HE \\
TCGA-DLBC & TCGA & 44 & Lymphoma & HE \\
TCGA-ESCA & TCGA & 158 & Esophagus & HE \\
TCGA-Frozen & TCGA & 18,394 & Pancancer & HE \\
TCGA-GBM & TCGA & 860 & Brain & HE \\
TCGA-HNSC & TCGA & 472 & Head and Neck & HE \\
TCGA-KICH & TCGA & 121 & Kidney & HE \\
TCGA-KIRC & TCGA & 519 & Kidney & HE \\
TCGA-KIRP & TCGA & 300 & Kidney & HE \\
TCGA-LGG & TCGA & 844 & Brain & HE \\
TCGA-LIHC & TCGA & 379 & Liver & HE \\
TCGA-LUAD & TCGA & 541 & Lung & HE \\
TCGA-LUSC & TCGA & 512 & Lung & HE \\
TCGA-MESO & TCGA & 87 & Mesothelioma & HE \\
TCGA-OV & TCGA & 107 & Ovaries & HE \\
TCGA-PAAD & TCGA & 209 & Pancreas & HE \\
TCGA-PCPG & TCGA & 196 & Pheochromocytoma and Paraganglioma & HE \\
TCGA-PRAD & TCGA & 449 & Prostate & HE \\
TCGA-READ & TCGA & 165 & Colon & HE \\
TCGA-SARC & TCGA & 600 & Sarcomas & HE \\
TCGA-SKCM & TCGA & 475 & Skin & HE \\
TCGA-STAD & TCGA & 442 & Stomach & HE \\
TCGA-TGCT & TCGA & 254 & Testicle & HE \\
TCGA-THCA & TCGA & 519 & Thyroid & HE \\
TCGA-THYM & TCGA & 181 & Thymoma & HE \\
TCGA-UCEC & TCGA & 566 & Uterus & HE \\
TCGA-UCS & TCGA & 91 & Uterus & HE \\
TCGA-UVM & TCGA & 80 & Melanoma & HE \\\midrule
TCGA & TCGA & 30,159 & Multi Organs & HE \\
\bottomrule
\end{tabular}
\end{table}

\begin{table}[t]
\centering
\small
\caption{The number of slides and processed patches of datasets used in this study. "-" represents the dataset only providing regions of interests (ROIs). }
\label{tab: whole_datasets}
\begin{tabular}{@{}lllllll@{}}
\toprule
\textbf{Dataset} & \textbf{Data Type} & \textbf{Number} & \textbf{Data Type} & \textbf{Number} & \textbf{Organ} & \textbf{Stain} \\ \midrule
TCGA \cite{weinstein2013cancer} & Slide & 30,159 & Patch & 120,496,200 & Multi Organs & HE \\
GTEx\_Portal \cite{carithers2015novel} & Slide & 25,711 & Patch & 31,892,017 & Multi Organs & HE \\
PANDA \cite{bulten2022artificial} & Slide & 10,616 & Patch & 905,206 & Prostate & HE \\
CPTAC \cite{edwards2015cptac} & Slide & 7,255 & Patch & 11,768,225 & Multi Organs & HE \\
DiagSet \cite{koziarski2024diagset} & Slide & 5,096 & Patch & 2,500,385 & Prostate & HE \\
ACROBATS2023 \cite{weitz2023multi} & Slide & 1,943 & Patch & 76,128 & Breast & HE and IHC \\
NADT-Prostate \cite{wilkinson2021nascent} & Slide & 1,404 & Patch & 919,847 & Prostate & HE and IHC \\
\bottomrule
\end{tabular}
\end{table}




% \end{appendices}


\end{document}